\documentclass[a4paper,fleqn]{cas-sc}

\usepackage[authoryear,longnamesfirst]{natbib}

\usepackage{comment}
\usepackage{amsmath,amsfonts}
\usepackage{algorithmic}
\usepackage{array}
\usepackage[labelformat=parens,labelsep=space]{subcaption}
\usepackage{textcomp}
\usepackage{stfloats}
\usepackage{url}

\usepackage{tabularray}
\usepackage[section]{placeins}
\usepackage{verbatim}
\usepackage{graphicx}
\usepackage{multirow}
\usepackage[inkscapelatex=false]{svg}
\usepackage{hyphenat}  
\usepackage{diagbox}
\usepackage{footnote}

\usepackage[nolist,nohyperlinks]{acronym}

\usepackage{cleveref}

\usepackage{silence}
\begin{document}

\begin{titlepage}
\begin{center}
\vspace*{1cm}

\textbf{\large Design Insights and Comparative Evaluation of a Hardware-Based Cooperative Perception Architecture for Lane Change Prediction}

\vspace{1.5cm}

Mohamed Manzour$^{1}$ (ahmed.manzour@uah.es), Catherine M. Elias$^{2}$ (catherine.elias@ieee.org), Omar M. Shehata$^{3}$ (omar.mohamad@guc.edu.eg), Rubén Izquierdo$^{1}$ (ruben.izquierdo@uah.es) and Miguel \'Angel Sotelo$^{1}$ (miguel.sotelo@uah.es) \\

\hspace{10pt}

\begin{flushleft}
\small  
$^{1}$ Computer Engineering Department, University of Alcalá, Madrid, Spain \\
$^{2}$ C-DRiVes Lab, Computer Science and Engineering Department, German University in Cairo, Egypt\\
$^{3}$ Multi-Robot Systems Lab, Mechatronics Engineering Department, German University in Cairo, Egypt\\

\vspace{1cm}
\textbf{Corresponding Author:} \\
Mohamed Manzour \\
San Diego Square, s/n. 28801, Alcalá de Henares, Madrid, Spain \\
Tel: +34 685687603 \\
Email: ahmed.manzour@uah.es

\end{flushleft}        
\end{center}
\end{titlepage}

\let\WriteBookmarks\relax
\def\floatpagepagefraction{1}
\def\textpagefraction{.001}
\shorttitle{Design Insights and Comparative Evaluation of a Hardware-Based Cooperative Perception Architecture for Lane Change Prediction}
\shortauthors{Mohamed Manzour et~al.}

\title[mode = title]{Design Insights and Comparative Evaluation of a Hardware-Based Cooperative Perception Architecture for Lane Change Prediction}                      

\tnotetext[1] {This research has been funded by the HEIDI project of the European Commission under Grant Agreement: 101069538.}

\author{Mohamed Manzour}[orcid=0009-0007-2009-3573]
\cormark[1]
\ead{ahmed.manzour@uah.es}

\author{Catherine M. Elias}[orcid=0000-0002-1444-9816]
\ead{catherine.elias@ieee.org}

\author{Omar M. Shehata}[orcid=0000-0002-3604-3534]
\ead{omar.mohamad@guc.edu.eg}

\author{Rubén Izquierdo}[orcid=0000-0002-6722-3036]
\ead{ruben.izquierdo@uah.es}

\author{Miguel {\'A}ngel Sotelo}[orcid=0000-0001-8809-2103]
\ead{miguel.sotelo@uah.es}

\cortext[cor1]{Corresponding author}

\affiliation{organization={Computer Engineering Department, University of Alcalá},
                state={Madrid},
                country={Spain}}

\affiliation{organization={ C-DRiVes Lab, Computer Science and Engineering Department, German University in Cairo},
                country={Egypt}}

\affiliation{organization={Multi-Robot Systems Lab, Mechatronics Engineering Department, German University in Cairo},
                country={Egypt}}

\begin{acronym}[TDMA] 

  \acro{ML}{Machine Learning}
  \acro{XGBoost}{eXtreme Gradient Boosting}
  \acro{NGSIM}{Next Generation Simulation}
  \acro{SVM}{Support Vector Machine}
  \acro{RF}{Random Forest}
  \acro{DBN}{Dynamic Bayesian Network}
  \acro{HMM}{Hidden Markov Model}
  \acro{DL}{Deep Learning}
  \acro{CNN}{Convolutional Neural Network}
  \acro{PREVENTION}{PREdiction of VEhicles iNTentIONs}
  \acro{TCN}{Temporal Convolutional Network}
  \acro{RNN}{Recurrent Neural Network} 
  \acro{LSTM}{Long Short-Term Memory} 
  \acro{GRU}{Gated Recurrent Unit}
  \acro{GNN}{Graph Neural Network}
  \acro{GCN}{Graph Convolutional Network}
  \acro{GAT}{Graph Attention Network}
  \acro{ST-GNN}{Spatio-Temporal Graph Neural Network}
  \acro{KGE}{Knowledge Graph Embedding}
  \acro{LiDAR}{Light Detection and Ranging}
  \acro{Radar}{Radio Detection and Ranging}
  \acro{IMU}{Inertial Measurement Unit}
  \acro{CAN}{Controller Area Network}
  \acro{NMPC}{Nonlinear Model Predictive Control}
  \acro{CRASH}{CARLA Risky-lane-change Anticipation in Simulated Highways}
  \acro{GPS}{Global Positioning System}
  \acro{GPU}{Graphics Processing Unit}
  \acro{IDM}{Intelligent Driver Model}
  \acro{MOBIL}{Minimizing Overall Braking Induced by Lane Changes}
  \acro{MPC}{Model Predictive Control}
  \acro{PID}{Proportional–Integral–Derivative}
  \acro{RSU}{Roadside Unit}
  \acro{FPS}{Frames Per Second}
  \acro{R-CNN}{Region-based Convolutional Neural Network}
  \acro{YOLO}{You Only Look Once}
   
  \acro{V2X}{Vehicle-to-Everything}
  
  \acro{THW}{Time Headway}
  \acro{TTC}{Time-To-Collision}

\end{acronym}

\begin{abstract}
Research on lane change prediction has gained attention in the last few years. Most existing works in this area have been conducted in simulation environments or with pre-recorded datasets, these works often rely on simplified assumptions about sensing, communication, and traffic behavior that do not always hold in practice. Real-world deployments of lane-change prediction systems are relatively rare, and when they are reported, the practical challenges, limitations, and lessons learned are often under-documented. This study explores cooperative lane-change prediction through a real hardware deployment in mixed traffic and shares the insights that emerged during implementation and testing. We highlight the practical challenges we faced, including bottlenecks, reliability issues, and operational constraints that shaped the behavior of the system. By documenting these experiences, the study provides guidance for others working on similar pipelines.
\end{abstract}


\maketitle

\section{Introduction}
\label{sec:introduction}
\begin{figure}[ht]
\centering
\includegraphics[width=\columnwidth]{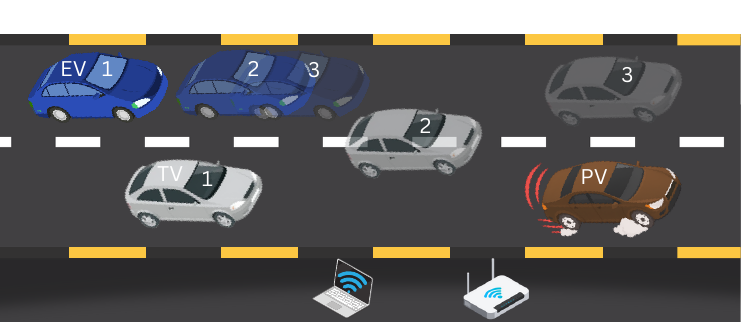}
\caption{Cooperative lane-change scenario. The Target Vehicle (TV) follows a Preceding Vehicle (PV) in the right lane, while the Ego Vehicle (EV) travels in the left lane. When the PV brakes, the TV shares its sensory data via a relay to the EV, enabling the EV to predict the TV's left lane change and create a safe gap.}
\label{fig:introduction_lane_changing_example}
\end{figure}

Traffic accidents remain a major global concern, with lane-change maneuvers recognized as one of the significant contributors to collision risk. Anticipating these maneuvers has become an important research focus, supporting both traffic safety and the safe integration of autonomous and assisted driving technologies. Over the past decade, numerous models have been developed for lane-change prediction. However, most existing works have been designed and validated using simulation environments or pre-recorded datasets. While these settings allow for benchmarking and controlled evaluation, they often rely on simplified assumptions about sensing, communication, and vehicle behavior that do not fully capture the complexity of real-world operation. Real-world deployments of lane-change prediction systems are relatively rare, and when they are reported, their practical challenges, limitations, and insights remain under-documented. To illustrate the setting more concretely, consider the left lane change scenario shown in \Cref{fig:introduction_lane_changing_example}. The Ego Vehicle (EV) is driving in the left lane, while the Target Vehicle (TV) is moving in the right lane behind a Preceding Vehicle (PV). When the PV suddenly brakes, the TV must change lanes to avoid a collision. The TV senses its surroundings using onboard sensors. However, instead of acting in isolation, it communicates its sensory data to an external relay unit. This relay processes the sensory data and transmits it to the EV in a form suitable for prediction. Using this information, the EV anticipates the TV’s left lane change maneuver and adapts its longitudinal behavior in time to create a safe gap for the lane change. Without this cooperative exchange, the EV would simply continue at its speed, leaving insufficient space and making the maneuver unsafe. This example illustrates how cooperation between vehicles, supported by an intermediate relay, enables timely and reliable lane-change prediction.

Building on this scenario, the present study examines a cooperative-perception pipeline deployed in mixed traffic, which contains human-driven, assisted, and autonomous vehicles. The aim is not to propose a new architecture but to share the insights and experiences gained during the real-world hardware deployment. We focus on documenting the bottlenecks, reliability issues, and operational constraints encountered across perception, prediction, and control. By reporting these findings, the study provides experience-based guidance that may support researchers and practitioners working on other similar systems.

\section{Literature and Research Foundation}
\label{sec:related-work}
\subsection{Literature Review}
\subsubsection{Dataset-based Validation}
Many early studies have employed traditional \ac{ML} models to predict the intentions of surrounding vehicles, primarily leveraging pre-recorded datasets without real-world or simulation-based validation. \ac{XGBoost} was applied to the \ac{NGSIM} dataset by \cite{zhang2022xgboost}, while \cite{syama2022ensemble} employed both \ac{SVM} and \ac{RF} on the same dataset. Logistic regression was adopted by \cite{ma2021lane} on the highD dataset. Across these works, evaluation was limited to dataset-based numerical validation, with no implementation in simulation environments or on physical hardware. All studies assumed seamless and error-free communication between vehicles, using datasets captured from a third-person perspective. As the highD dataset was recorded from a drone-mounted camera, and the NGSIM dataset a camera installed on fixed roadside infrastructure, both providing complete and noise-free vehicle state information which are conditions that does not reflect the challenges of real-world perception and communication systems.

Similarly, some studies relied on the Bayesian network family of models, which includes Static Bayesian Networks, \acp{DBN}, \acp{HMM}, and Naive Bayes classifiers. For example, \cite{li2019dynamic} and \cite{liu2024peril} applied \acp{DBN} to the NGSIM and highD datasets, respectively. Works \cite{xia2021human,li2023early} employed \acp{HMM}, relying on kinematic inputs such as velocity, acceleration, and relative position obtained from utilizing the NGSIM dataset.

Other studies have relied on convolutional \ac{DL} models, using image-based inputs rather than numerical features. Works \cite{izquierdo2019experimental,liang2022lane,fernandez2020two,biparva2022video,izquierdo2021vehicle} employed variants of \acp{CNN} and hybrid CNN–LSTM architectures on the \ac{PREVENTION} dataset, which provides an ego-vehicle perspective of the driving scene. These implementations varied in their architectural choices, including standard \acp{CNN}, 3D \acp{CNN}, and spatio-temporal ConvNets for capturing both spatial and temporal dependencies in the image sequences.

Building on these spatio-temporal approaches, the research focus gradually shifted toward sequential \ac{DL} models that emphasize the temporal dynamics of driving behavior. These models include \acp{RNN}, \ac{LSTM} networks, and \acp{GRU}, as well as their advanced variants such as bidirectional \acp{LSTM} and attention-based \acp{LSTM}. For instance, \cite{su2018learning} utilized an \ac{LSTM} model trained on the \ac{NGSIM} dataset, while work \cite{amer2024enhancing} applied \acp{LSTM} to the \ac{PREVENTION} dataset, both works relied on kinematic inputs such as vehicle positions, accelerations, and relative velocities. Work \cite{shangguan2022proactive} trained an \ac{LSTM} model on the highD dataset, while works \cite{scheel2019attention} and \cite{scheel2022recurrent} incorporated attention mechanisms into their \ac{LSTM} architectures. Additionally, \cite{laimona2020implementation} conducted a comparative study between \ac{RNN} and \ac{LSTM} models using the \ac{PREVENTION} dataset, where the input consisted of sequences of X-Y centroid coordinates extracted from vehicle detection outputs.

More recent studies used transformers as their prediction model. For example, \cite{gao2023dual,guo2025vehicle,lu2025knowledge} employed the \ac{NGSIM} and highD datasets together. These works generally use numerical inputs such as relative distances and velocities of surrounding vehicles, along with the lateral and longitudinal positions of the EV. In contrast, \cite{peng2025lc} employed Large Language Models (LLMs) to predict both lane-change intentions and lane-change trajectories. The model was provided with a prompt containing vehicle attributes such as position, velocity, type, and lane ID, and further leveraged a Chain-Of-Thought (CoT) mechanism to produce interpretable explanations for its predictions.

\acp{GNN} were also explored. For instance, \cite{lu2025lane} employed \acp{GAT} and \acp{GCN}, while \cite{li2023social} adopted \acp{ST-GNN}. Both approaches were trained and evaluated on the NGSIM and highD datasets.

Other works have explored graph representation learning approaches, particularly \acp{KGE}. For example, \cite{manzour2024vehicle,hussien2025rag} employed the highD dataset and instead of relying on numerical features, they utilized linguistic input formats. This formulation enhanced the interpretability of the models. Then, Bayesian inference was applied as a downstream task on top of the learned embeddings.

Some studies did not rely on publicly available datasets such as NGSIM or highD, but instead collected private datasets to validate their architectures or models. Most of these works focused primarily on predicting the intentions of the EV rather than surrounding vehicles. For instance, Papers \cite{xing2020ensemble,jain2015car} collected their perception data using cameras, GPS, and CAN bus signals. Other works \cite{berndt2008continuous,li2016lane,li2024lane,rastin2024multi} relied on multi-sensor setups by using different combination of sensors such as cameras, \ac{Radar}, \ac{LiDAR}, HD-maps, \ac{IMU}, and \ac{CAN} bus information to capture a richer description of vehicle dynamics and its surroundings.

From these studies, we see that dataset-based validation has mostly been limited to extracting perception information (often simplified) and applying prediction models, without extending to decision-making or control. No coupling was made with even basic longitudinal actions such as braking or smooth deceleration. Furthermore, none of the works progressed toward simulation-based validation or real-world hardware testing, which limits their relevance and applicability in actual driving environments.
\subsubsection{Simulation-based Validation}
Some studies did not limit their validation to datasets alone but extended it to simulation-based evaluations combined with decision-making modules. For example, \cite{li2021lane} reported numerical validation on the NGSIM dataset and further tested the model in the CarSim environment. The work indicated that the planning stage was based on a \ac{NMPC} scheme, but did not specify whether this planning was actually executed by the EV, it may have remained at the planning level without execution. Similarly, \cite{manzour2025explainable} validated its approach on the highD and \ac{CRASH} datasets, aiming to predict both risky and safe lane-change maneuvers. Their work was extended in the CARLA simulator, where their planing was executed only in the longitudinal direction (brake or smooth deceleration), and no details were provided on the specific controller used for execution.
\subsubsection{Hardware-based Validation}
A smaller set of works carried this further into hardware real-world testing. For instance, \cite{morris2011lane,doshi2011road} validated their prediction models in on-road conditions. However, these efforts focused solely on the prediction stage without incorporating any planning or control modules. In contrast, other studies combined real-world testing with decision-making. Work \cite{manzour2025real} implemented planning in the longitudinal direction only, while \cite{gonzalo2022testing} addressed both lateral and longitudinal planning. Nevertheless, in both cases, the specific controllers employed to execute these planned maneuvers were not explicitly described. Also, \cite{manzour2025real} introduced communication between vehicles, where sensory data from one vehicle was processed and transmitted to another, enabling cooperative maneuver anticipation.

\subsection{Observed Functional Blocks}
From the reviewed literature, a pattern of functional building blocks emerges. 
Dataset-based works focused mainly on extracting vehicle states from sensors or datasets (Perception) and applying models to anticipate maneuvers (Prediction). 
Later studies extended these models into simulation, where predicted intentions were connected to Planning and Control. Finally, hardware-based implementations introduced real-world execution, and in some cases Communication between vehicles, where information was exchanged to enable cooperative behavior. Together, these observations suggest that most lane-change prediction systems can be described in terms of four interconnected blocks: Perception, Communication, Prediction, and Planning \& Control.

\subsection{General Prediction Pipeline}
\begin{figure}[!htbp]
\centering
\includegraphics[width=\columnwidth]{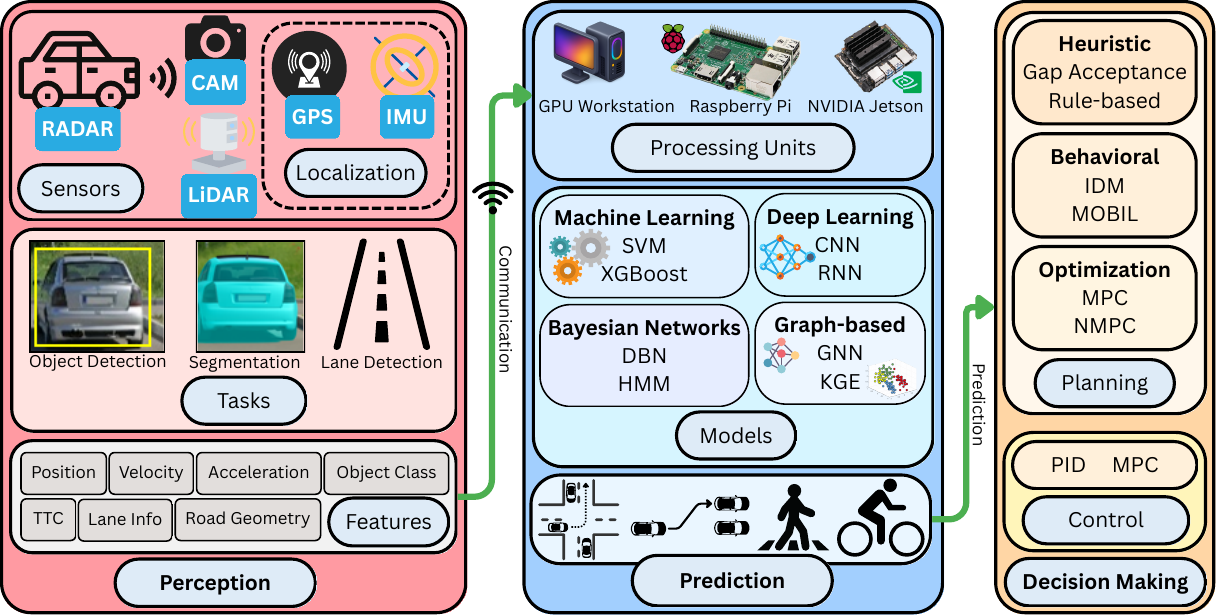}
\caption{Generalized prediction pipeline covering perception, prediction, and decision-making with control, applicable to diverse scenarios and agents in autonomous and cooperative systems.}
\label{fig:prediction_pipeline}
\end{figure}

Based on these observations, a generalized prediction pipeline can be formulated. As illustrated in \Cref{fig:prediction_pipeline}, the process begins with the
Perception module, where raw sensory inputs such as cameras, \ac{LiDAR}, \ac{Radar}, \ac{GPS}, or \ac{IMU} are transformed into meaningful representations. These include tasks such as object detection, lane detection, segmentation, or 3D point cloud generation. From these outputs, relevant features are extracted, including vehicle position, velocity, acceleration, \ac{TTC}, surrounding object classes, and lane geometry. When cooperation is considered, the Communication module enables these features to be shared between vehicles or with roadside infrastructure using, for example, Wi-Fi links or roadside units. The Prediction module then builds on this shared information to anticipate the future behavior of observed agents. Depending on the computational platform (ranging from high-performance \acp{GPU} to embedded devices such as Raspberry Pi and NVIDIA boards) different algorithmic families can be employed, including \ac{ML} (e.g., \ac{SVM}, \ac{XGBoost}), \ac{DL} (e.g., \acp{CNN}, \acp{RNN}, Transformers), and graph-based approaches (e.g., \acp{GNN}, \acp{KGE}). Such models are capable of predicting not only vehicle trajectories and lane-change maneuvers but also the intentions of other road users, including pedestrians and cyclists, in complex traffic environments. The resulting predictions are then passed to the Decision-Making and Control modules, which plan the EV’s longitudinal and lateral actions. This planning may rely on behavioral models such as \ac{IDM} or \ac{MOBIL}, heuristic or rule-based strategies such as gap acceptance, or optimization-based methods like \ac{MPC} or \ac{NMPC}. Finally, controllers such as \ac{PID} or \ac{MPC} ensure that these planned actions are carried out reliably, enabling safe and coordinated operation.
A similar end-to-end architecture has also been demonstrated on real hardware in previous works. For example, \cite{manzour2025real} focused on lane-change prediction, while \cite{manzour2024development} addressed pedestrian crossing prediction. Both studies followed a similar architectural flow as described here, starting from perception, then communication and prediction, and ending with decision-making with control.
This generality highlights the pipeline as a unifying framework for many prediction tasks in autonomous and cooperative systems. However, it should be emphasized that in this study, the focus remains specifically on lane-change intention prediction.

\subsection{Study Objective}
The objective of this work to study how a cooperative prediction pipeline behaves when deployed in hardware and to share the insights gained from this process. The focus is on documenting what was observed in practice rather than presenting a novel method. 
In particular, this study aims to:
\begin{itemize}
    \item Examine the operation of each module in the pipeline (Perception, Communication, Prediction, Planning \& Control) under real-world conditions.  
    \item Report the practical challenges and bottlenecks encountered in each module, such as sensing limitations, computational load, communication delays, and thermal issues.  
    \item Compare different processing alternatives to highlight trade-offs in performance and feasibility.  
    \item Share the lessons learned during implementation, providing experience-based guidance for researchers and practitioners working on similar cooperative prediction systems.  
\end{itemize}

\section{Experimental Setup}\label{sec:experimrntal_setup}
\begin{figure*}[ht]
\centering
\includegraphics[width=\linewidth]{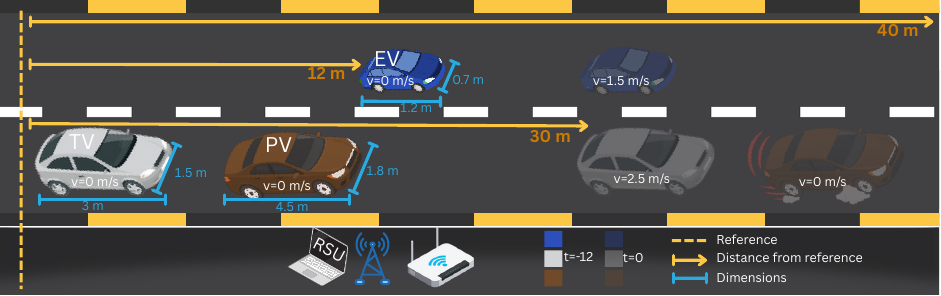}
\caption{Experimental scenario with three vehicles: EV, TV, and the PV. At $t=-12$ seconds, the vehicles are shown in solid color as they move along the 40-meter test track. As time advances, transparent overlays represent their updated positions until $t=0$, which is the moment when the PV brakes suddenly and the TV initiates a left lane change.}
\label{fig:experiment_setup}
\end{figure*}

\Cref{fig:experiment_setup} illustrates the setup of the
studied scenario, which involves three vehicles: the EV, the TV, and the PV. The PV is a human-driven Peugeot 301 measuring 4.5 meters in length and 1.8 meters in width. Its role in the scenario is to move ahead of the TV and then apply a sudden brake, creating a risk that forces the TV to react. The TV is a golf cart equipped with sensors, with dimensions of 3.0 x 1.5 meters, and is also manually driven. It perceives the environment and responds to the PV’s braking by attempting a left-lane change. The EV is a scaled-down vehicle, 1.2 meters long and 0.7 meters wide, which does not carry its own perception sensors. Instead, it is fitted with actuation and prediction modules. Initially, all three vehicles start from stationary positions on a straight 40-meter test track. The EV is positioned 12 meters ahead, but since its average speed is only 1.5 m/s compared to 2.5 m/s for the TV and PV, it is gradually overtaken. After approximately 12 seconds, the PV brakes suddenly, forcing the TV to attempt a lane change to avoid collision. Two experimental cases were considered: in the first, the EV has no prediction module and continues straight without anticipating the TV’s maneuver, leaving the TV with no safe gap and forcing it into harsh braking. In the second case, the EV is equipped with the prediction module. Here, the TV sends its perceptual data to a relay device, which processes and forwards it to the EV. Using this cooperative information, the EV anticipates the TV’s lane-change intention, decelerates in advance, and creates the necessary space for the maneuver. Figure 3 shows the positions of the vehicles at two reference times. At $t=-12$ seconds, when the vehicles are still in steady motion before the PV applies the brake, the vehicles are shown in solid color. As time progresses toward $t=0$, transparent overlays are used to indicate the advancing positions of the vehicles showing the critical moment when the PV brakes suddenly and the TV initiates a left lane change.

\begin{figure}[ht]
\centering
\includegraphics[width=\linewidth]{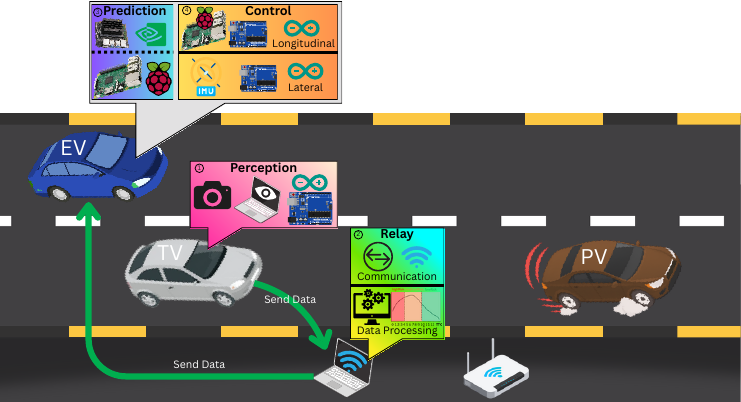}
\caption{Cooperative lane-change prediction scenario illustrated through the hardware components used in the experiment. The TV is equipped with a ZED camera, an onboard laptop, and a microcontroller to sense the environment. This perceptual data is transmitted to a relay Roadside Unit (RSU), which processes and communicates the information to the EV. The EV hosts the prediction module on Nvidia Jetson Nano board and executes control through a Raspberry Pi–microcontroller chain for longitudinal actuation, and a microcontroller for lateral lane-keeping. Through this cooperation, the EV anticipates the TV’s maneuver and adjusts its longitudinal speed to ensure safe interaction.}
\label{fig:hardware_components}
\end{figure}

\begin{figure*}[ht]
\centering
\includegraphics[width=\linewidth]{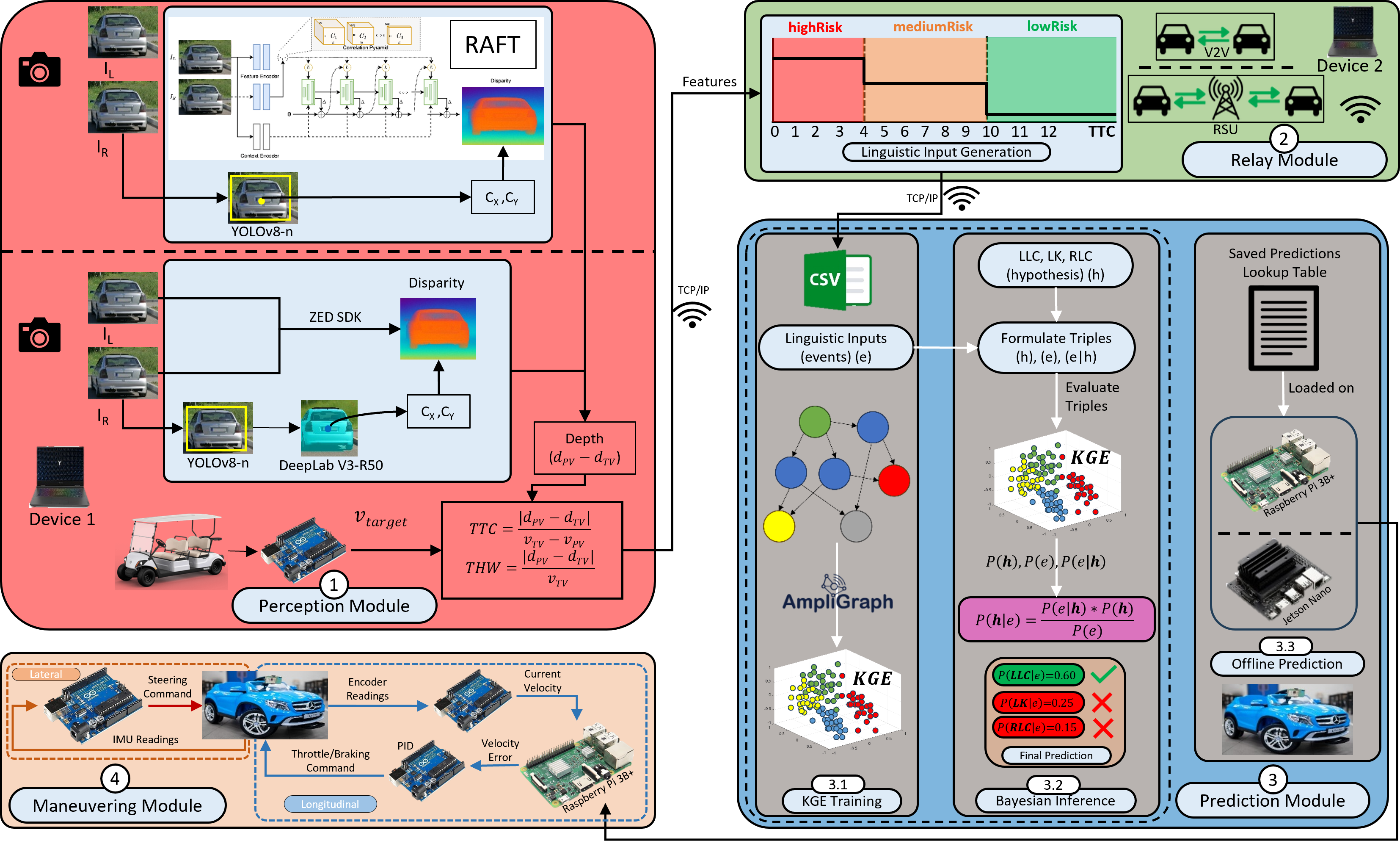}
\caption{Functional workflow of the prediction pipeline, showing the studied methods applied in perception, relay, prediction, and maneuvering modules.}
\label{fig:methodology}
\end{figure*}

\Cref{fig:hardware_components} provides an overview of the hardware components integrated into each vehicle and the functional modules they support.
In the TV, a ZED stereo camera captures the environment, and its outputs are processed by an onboard laptop \textit{(Device 1)} that executes different perception algorithms. A microcontroller is also installed to measure the TV’s own velocity. This perceptual information is then sent to the relay module, implemented as a \ac{RSU} \textit{(Device 2)}, which is responsible for receiving, processing, and transmitting the data to the EV. The EV contains the prediction module, which was tested across different computing platforms including a standard laptop, a Raspberry Pi, and an NVIDIA Jetson Nano board. For control, the EV separates longitudinal and lateral functions. Longitudinal actuation (throttle and braking) is implemented through a Raspberry Pi that serves as a high-level device, receiving inputs from the prediction module and the EV’s onboard encoder, and sending throttle/braking commands to another low level controller. Lateral control (responsible for lane-keeping) is handled by a microcontroller that reads IMU data and sends steering commands to the vehicle. The PV, on the other hand, is a human-driven vehicle used only to create the risky scenario by braking suddenly and does not incorporate additional sensing or communication modules. \Cref{fig:experiment_setup} and \Cref{fig:hardware_components} therefore illustrates the scenario flow and how each vehicle is set up to fulfill a distinct role in the cooperative experiment: the PV as the disturbance, the TV as the sensor-equipped agent that will make the lane change, and the EV as the cooperative partner equipped with prediction and control. \Cref{fig:methodology} illustrates the studied methods and tasks within each module. The perception module handles sensing operations such as object detection, segmentation, and the estimation of \ac{TTC} and \ac{THW} between the TV and the PV. The relay module acts as an intermediate processing and communication unit, converting numerical values into linguistic categories and ensuring a standardized format for data transmission to the prediction module. The prediction module integrates \ac{KGE} with Bayesian inference to anticipate the TV’s maneuver, supported by an offline implementation for real-time operation on embedded platforms. Finally, the maneuvering module translates these predictions into vehicle actions: longitudinal braking and throttle commands are executed through a Raspberry Pi–Arduino chain, while lateral steering is coordinated through a low level microcontroller. Together, these studied architecture demonstrate how sensing, communication, prediction, and actuation are connected across the cooperative pipeline. The following subsections detail the operation of each module under real-world conditions.

\section{Perception and Localization Modules}
\subsection{Target Vehicle System Breakdown}
In a regular Autonomous Driving Stack (ADS), the vehicle is equipped with the localization, perception, mapping, planning, control, and system integration. However, in this work, the focus is to investigate the insights and challenges associated with incorporating the Cooperative Perception (CP) and how it will affect the quality of the risky maneuver prediction rather than presenting a full ADS. Therefore, the utilized TV is designed to be equipped with 2 main modules from the ADS. The first module is the localization module that is responsible for having information about the vehicle states. This information is necessary for the second module, which is perception. This module is mainly responsible for two main tasks; Object Detection and Safety Features Extraction. In this section, an experiment setup of the TV will be demonstrated as well as the implementation of the perception module and the message prepared to be sent to the relay.

\subsection{Target Vehicle Setup}
\begin{figure}[ht]
\centering
\includegraphics[width=0.7\linewidth]{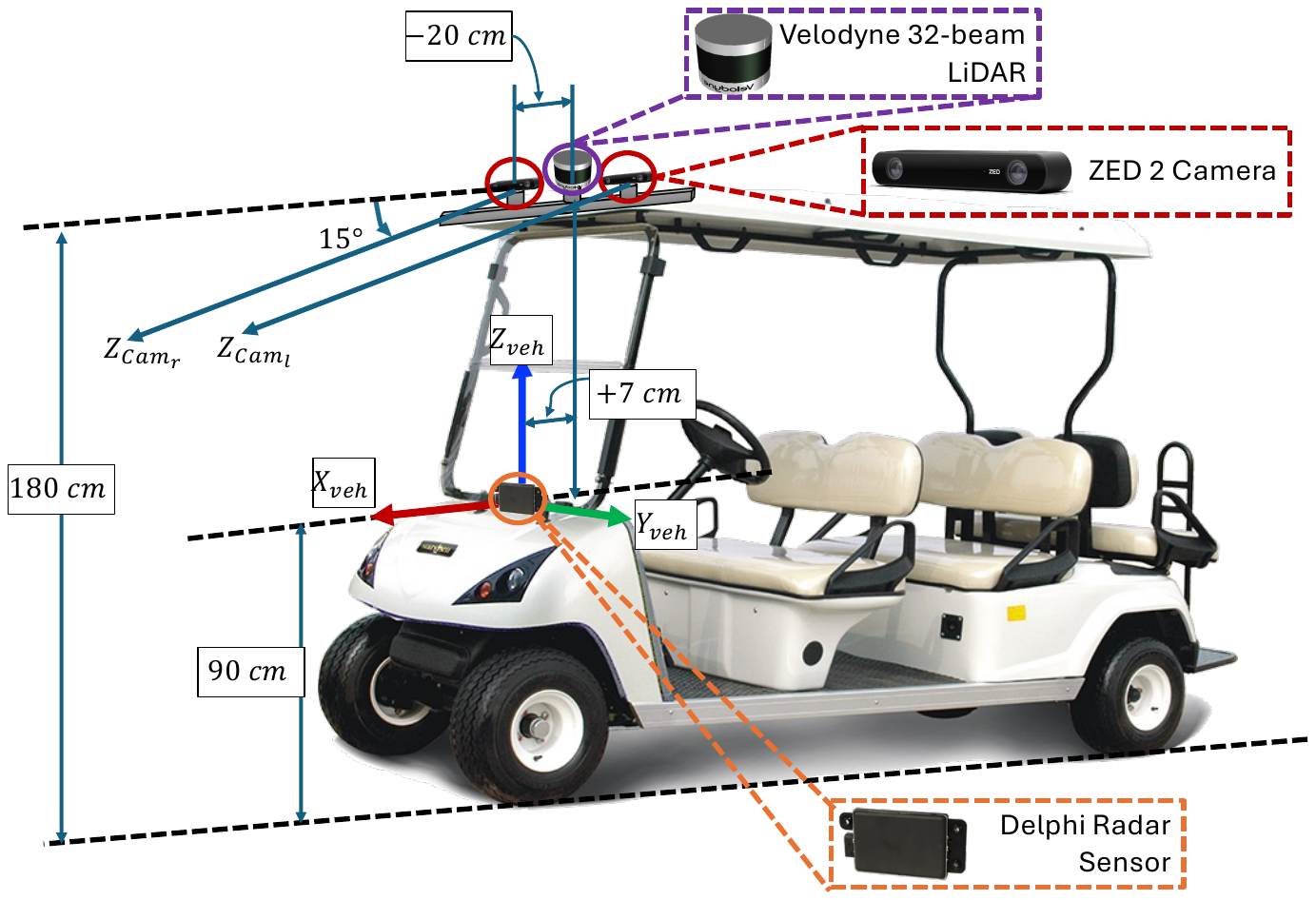}
\caption{The TV Sensors Mounting}
\label{fig:VT_Setup}
\end{figure}

During the example pipeline, the TV used is a golf cart of type Marshell Utility Electric Golf Cart DG-C4+2 shown in \Cref{fig:VT_Setup}.
The TV is equipped with a number of sensors and processors that are used to perform the localization and perception modules. These sensors are mounted in the vehicle as shown in \Cref{fig:VT_Setup}. The added perception sensors are:
\begin{itemize}
    \item 2 Stereolabs ZED 2 camera with a resolution of $672 \times 376 px @ 10 \ac{FPS}$ (neural depth mode) and $119.89 mm$ baseline. The 2 cameras are mounted on the vehicle roof on both sides with a spacing of $\pm 20 cm$ from the center of the vehicle and on a distance of $180 cm$ measured from the ground.
    \item The vehicle is also equipped with 32-beam Velodyne LiDAR and Delphi radar sensors. However, in this work, visual perception will be the main method implemented. Yet, the TV platform is designed to test other different perception pipelines/algorithms. Therefore, these essential sensors are added.
    \item The perception is done on two separate laptops which are used to run each studied pipeline solely. The specs of these laptops are: a laptop with Nvidia GeForce RTX 4060 \ac{GPU}, and a laptop with NVIDIA GeForce RTX 3050 \ac{GPU}.
\end{itemize}

\subsection{Localization Module}
\begin{figure}[ht]
    \centering
    \includegraphics[width=0.7\linewidth]{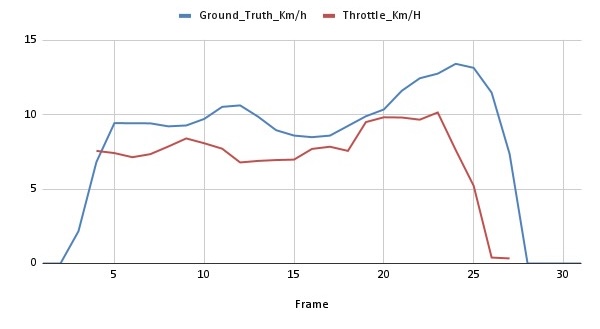}
    \caption{TV's Speed Profile Obtained from Throttle-calibrated Measurement Vs. Mobile GPS Sensor}
    \label{fig:Speed}
    \end{figure}

The localization module in the regular ADS is used mainly to get the TV's states including its position $P_{TV}$ using Global Navigation Satellite System (GNSS) sensors either globally or locally, the vehicle’s orientation  $Yaw_{TV}$ using IMU sensor, and some other important states customized to the work such as the longitudinal velocity $V_{TV}$ using encoders and acceleration $A_{TV}$ using accelerometers. In this work, since it is decided to disable the planning and control of the TV and only focus on the perception, it is only necessary to measure the vehicle’s longitudinal velocity $V_{{TV}_t}$ at each time sample $t$. This velocity is needed to compensate the relative measurement of the perception module. It is also important to mention that the time sample $t$ is determined based on the perception module rate. The velocity of the vehicle is obtained in this work using direct measurement from the vehicle’ throttle with calibration using the gear state. This velocity measurement is essential in estimating the absolute velocity of the PV $V_{PV_{Abs}}$. This measured TV's velocity is then compared with the velocity obtained from a mobile phone GPS. As shown in \Cref{fig:Speed}, the maximum absolute error is $3,\text{km/h}$ in the absence of rapid speed changes. It can also be observed that the throttle measurement responds faster to speed changes compared to the velocity calculated by the GPS. Although the error in throttle-based measurement is relatively high, the throttle method was chosen for the rest of the experiments. This decision is mainly due to the fact that GPS readings can be faulty and are strongly affected by urban environments and poor signal quality. A closer inspection of the experimental site revealed the presence of a high-voltage station emitting strong electromagnetic fields, which can significantly degrade GPS accuracy. In addition, the experiments were conducted on a university campus surrounded by tall buildings, further weakening the GPS signal. While the throttle method is somewhat sensitive to the vehicle’s battery charging level, it remains the most suitable option, particularly because it provides more responsive measurements at a higher rate compared to GPS, which only outputs readings at a $1$-second interval.

\subsection{Perception Module}
For the implemented perception module, a layered pipeline composed of number of sequential stages has been structured with four consecutive stages. And as an initial step, the layered pipeline initializes the stereo camera, extracts, and stores its intrinsic parameters, such as the baseline, focal length, and principal points.

\subsubsection{Stage 1: Object Detection, Classification, \& Tracking}
\begin{figure}[ht]
  \centering
  \includegraphics[width=0.60\textwidth]{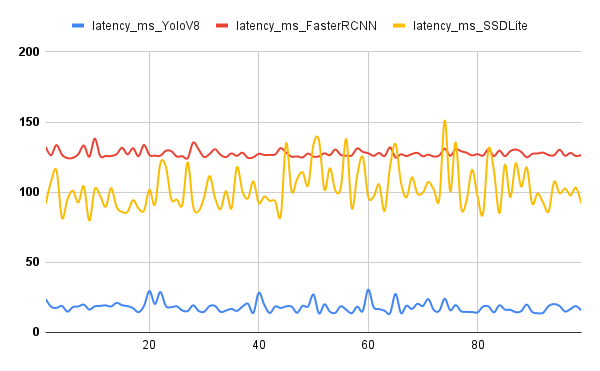}
  \fbox{\parbox{0.6\textwidth}{\centering
    \small Latency (ms) per frame for \ac{YOLO}v8\_n, Faster \ac{R-CNN}, and SSDLite320}}
  \caption{Frame‐by‐frame inference latency for each detector.}
  \label{fig:detection_latency}
\end{figure}

In this stage, all surrounding objects are detected, bounded in 2D boxes, and classified with one of the trained classes. In this work, the architecture is only trained to limited classes relevant to the architecture, namely the "\textit{car}" class. Furthermore, information about the class scores and persistent ID is added as output from the tracking. In the implementation of this stage, three different benchmark methods have been investigated. These methods are \ac{YOLO}v8-n \cite{ultralytics2025yolov8}, Faster \ac{R-CNN} \cite{girshick2015fast}, and SSDLite (a lightweight version of the Single Shot MultiBox Detector, SSD) \cite{howard2019searching}. To quantitatively assess the detector, the three methods have been examined on over 100 frames from a ZED SVO, logging per-frame latency, number of detections, and mean confidence. In terms of latency and throughput, as shown in \Cref{fig:detection_latency}, \ac{YOLO}v8-n achieves the lowest average inference latency of $\approx 18.4\,\text{ms}$ ($\approx 54$ \ac{FPS}), comfortably above our 10 \ac{FPS} real-time target. In contrast, Faster \ac{R-CNN} runs at $\approx 7.8$ \ac{FPS} and SSDLite at $\approx 9.9$ \ac{FPS}, making them marginal or unsuitable without further acceleration.
\Cref{tab:detection_stats} summarizes the obtained results to prove the superiority of \ac{YOLO}v8-n, especially in terms of the latency which is critical in our full cooperative system scenario. In addition, \ac{YOLO}v8-n achieves an average confidence of $0.70$, which is significantly more reliable than Faster \ac{R-CNN} ($0.22$) and SSDLite ($0.07$).
\begin{table}[ht]
  \centering
  \caption{Detection Benchmark Summary (100-frame mean ± std).}
  \label{tab:detection_stats}
  \begin{tabular}{lccc}
    \toprule
    \textbf{Model}     & \textbf{Latency (ms)} & \textbf{Avg.\ conf.} \\
    \midrule
    \ac{YOLO}v8-n          & $18.4 \pm 3.5$      & $0.70 \pm 0.05$ \\
    Faster \ac{R-CNN}       & $127.8 \pm 3.7$    & $0.22 \pm 0.03$ \\
    SSDLite320         & $101.2 \pm 12.1$   & $0.07 \pm 0.01$ \\
    \bottomrule
  \end{tabular}
\end{table}
Overall, \ac{YOLO}v8-n achieves the lowest latency, the most realistic number of detections, and the highest confidence, making it the most suitable choice for our system. So, it is decided to continue with \ac{YOLO}v-8 in ByteTrack \cite{zhang2022bytetrackmultiobjecttrackingassociating} mode.

\subsubsection{Stage 2: Depth Estimation} \label{sec:stage2}
The aim of this stage is to estimate the depth of the detected vehicles (the PV) relative to the TV. For this stage, two different state-of-the-art pipelines have been investigated.

\paragraph{\textbf{\textit{Pipeline 1: RAFT-Stereo Depth Estimation}}}
The first investigated method is Recurrent All-Pairs Field Transforms (RAFT)-Stereo \cite{lipson2021raft}. RAFT works directly with the left and right images from the stereo camera. It takes both images as input and runs for 24 iterations to produce a high resolution disparity map $d$. This disparity map is comprised of point pairs $d(u,v)$, where each value corresponds to the shift between the left and right images. This disparity is then converted to depth map $Z$ in which each point depth $Z(u,v)$ is computed using the below formula:
\begin{equation}\label{eq:disp}
Z(u,v) = \frac{fB}{d(u,v)}
\end{equation}
where $f$ is the focal length of the camera and $B$ is its baseline. This depth $Z(u,v)$ represents the distance between the camera frame and the point in the disparity map $d(u,v)$. The obtained point depth is further calibrated to compensate the $15^\circ$ tilt of the camera mounted on the vehicle and illustrated in \Cref{fig:VT_Setup}. The computed depth map points values are then passed through several filters to smooth the returned values including median filtering, Gaussian smoothing, and Bilateral filtering for edge preservation. After estimating the depth of the PV's center point $(u,v)$, the final vehicle's coordinates relative to the camera frame has been formulated:
\begin{equation}
    X = \frac{(u - c_x)\,Z}{f_x}, \quad
Y = \frac{(v - c_y)\,Z}{f_y}, \quad
Z = \text{depth}.
\end{equation}
where $(c_x, c_y)$ are the principle point offsets and $(f_x, f_y)$ are the camera's focal lengths. Finally, these coordinates are then translated from the camera frame to the vehicle frame to represent the position of the PV relative to the TV.

\begin{figure}[ht]
  \centering
  \begin{subfigure}[t]{0.24\textwidth}
    \includegraphics[width=\textwidth]{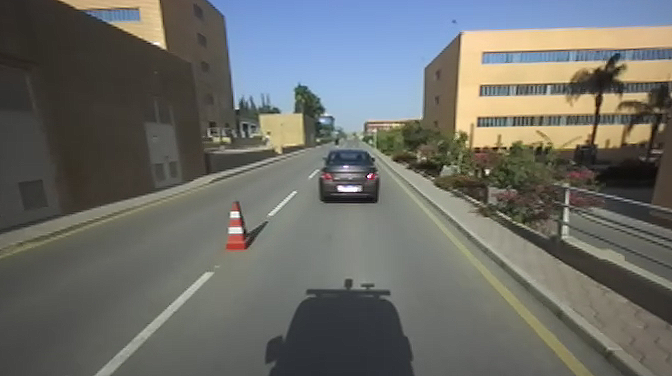}
    \caption{Input Frame}
    \label{fig:seg_a}
  \end{subfigure}
  \hfill
  \begin{subfigure}[t]{0.24\textwidth}
    \includegraphics[width=\textwidth]{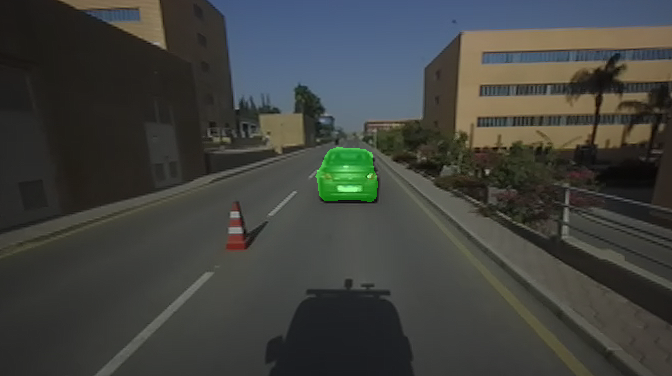}
    \caption{DeepLabV3‐ResNet50}
    \label{fig:seg_b}
  \end{subfigure}
  \begin{subfigure}[t]{0.24\textwidth}
    \includegraphics[width=\textwidth]{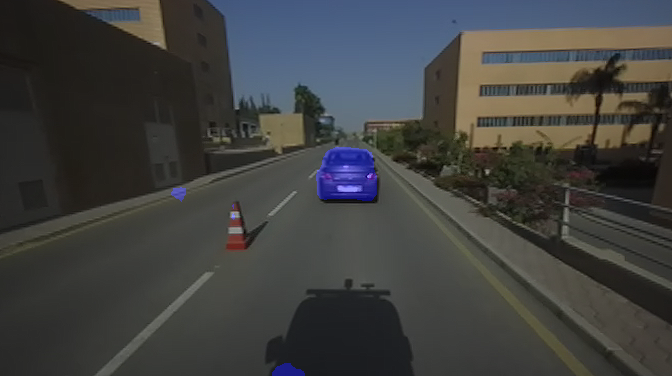}
    \caption{FCN‐ResNet50}
    \label{fig:seg_c}
  \end{subfigure}
  \hfill
  \begin{subfigure}[t]{0.24\textwidth}
    \includegraphics[width=\textwidth]{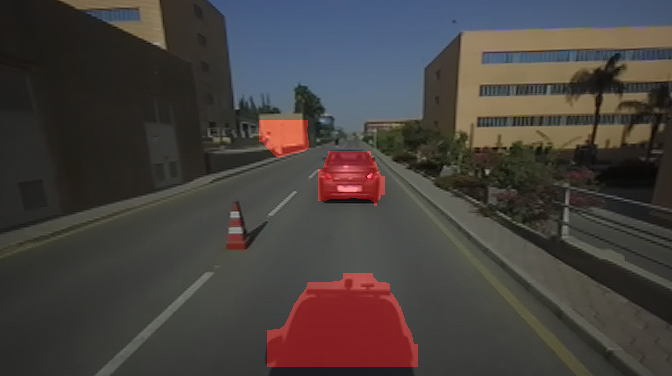}
    \caption{\ac{YOLO}v8s‐Seg}
    \label{fig:seg_d}
  \end{subfigure}
  \caption{Qualitative ROI‐based segmentation outputs for the three backbones.}
  \label{fig:segmentation_examples}
\end{figure}

\begin{figure}[ht]
  \centering
  \includegraphics[width=0.65\textwidth]{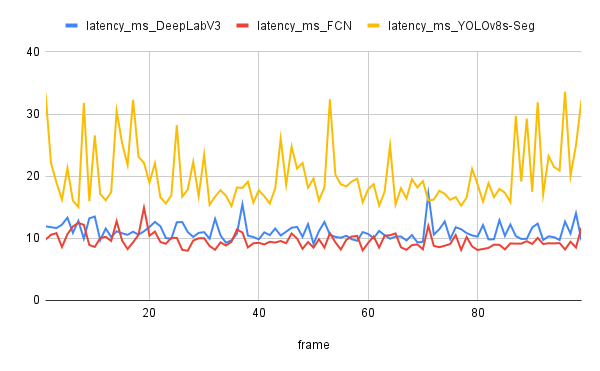}
  \caption{Frame‐by‐frame inference latency for DeepLabV3, FCN, and \ac{YOLO}v8s‐Seg (100 frames).}
  \label{fig:seg_latency}
\end{figure}

\paragraph{\textbf{\textit{Pipeline 2: ROI-Based Segmentation Depth Estimation}}}
For the second pipeline implementation, it is decided to work with segmented objects rather than just using bounding box as detection output. The main aim is to reduce the per-frame pixel throughput with the target of increasing the perception rate. Each detected box defines a Region-of-Interest (ROI), which is resized to the segmentation model’s input (e.g.\ $512 \times 512 px$). By focusing only on vehicle ROIs, the per‐frame pixel throughput is reduced by up to $85\%$. Three different backbones are compared and evaluated which are DeepLabV3 with Residual Neural Network (ResNet)-50 and ResNet-101 encoders \cite{Cite22}, Fully Convolutional Network (FCN)-ResNet-50/101 \cite{Cite21}, and \ac{YOLO}v8s‐Seg variant \cite{ultralytics2023yolov8}. DeepLabV3 incurs an average latency of approximately $11.2 \pm 1.1$ ms per ROI, FCN is the fastest at roughly $9.6 \pm 0.9$ ms, and the integrated \ac{YOLO}v8s‐Seg head runs at about $20 \pm 4.3$ ms. These results guide our choice of segmentation backbone under different real‐time constraints: FCN for maximum throughput, DeepLabV3 for highest mask quality, and \ac{YOLO}v8s‐Seg when unified inference is preferred. The investigated segmentation models are evaluated based on qualitative mask quality and the per-frame latency on 100 frames of ZED 2 left-view video. Each model outputs a binary mask per ROI, which is threshold and upsampled to the original box size. Then, the per‐model IoU and segmentation latency are measured to guide our selection under the 10–30 \ac{FPS} constraint. \Cref{fig:segmentation_examples} presents a side‐by‐side comparison of the three segmentation backbones on a representative vehicle ROI. It can be observed that the green DeepLabV3 mask offers the most complete coverage of the vehicle’s silhouette, effectively capturing the full extent of the car body. In contrast, the blue FCN mask adheres more closely to the car’s geometric outline but suffers from over‐segmentation of non‐vehicle elements (e.g., scene clutter), leading to false positives in the background. Finally, the red \ac{YOLO}v8s‐Seg mask shows irregular, “jagged” contours and even misclassifies the vehicle’s shadow and nearby wall as part of the vehicle, indicating both under‐ and over‐segmentation artifacts. The per-frame inference time for each segmentation model was measured over 100 ZED frames, and the latency traces are shown in \Cref{fig:seg_latency}.

Based on these results, it is decided to proceed with the DeepLabV3-ResNet50 semantic segmentation model \cite{chen2017rethinking} to produce a binary mask for cars, and overlays the mask. Once the PV is being detected and segmented, its depth is being estimated however in this method, the disparity map $d$ is obtained from the ZED’s Software Development Kit (SDK) that computes $d$ by matching each left-pixel to its counterpart in the right image. Afterwards using Equation \eqref{eq:disp}, the vehicle depth $Z$ is computed.

\subsubsection{Stage 3: Object Tracking \& Kinematic Estimation}
This stage is important to track the historical kinematical estimations of the object. This helps in extracting more kinematical properties as the object speed and yaw angle. An object tracker class is created to store the history of each object detected for a few frames, even after its disappearance from the camera view, to enhance the tracking of the objects. For this work, the most critical kinematical feature to be extracted is the PV speed, which is crucial for the safety features extraction. Two different trackers and kinematical estimation techniques are used depending on the implemented pipeline discussed in \Cref{sec:stage2}. For Pipeline 1, after computing the PV depth, its speed is calculated using the saved historical kinematical information as follows:
\begin{equation}
    v_{\mathrm{obj}}
  = - \frac{v_x\,X + v_z\,Z}{\sqrt{X^2 + Z^2}}
\end{equation}
where \(v_x\) and \(v_z\) is the change of the PV's position in relation to the time in the $x$ and $z$ directions respectively, while the $X$ and $Z$ are its coordinates in the 3D camera frame. 

For Pipeline 2, the implemented tracker class tracks per-object depth and velocity with Exponential Moving Average (EMA) smoothing technique. For each track, it maintains a rolling window of recent depth measurements and computes a smoothed depth as the mean over that window. An anchor-based velocity is calculated across the window, then an exponential moving average with coefficient \(\alpha=0.3\) at 20 \ac{FPS} is applied.

\subsubsection{Stage 4: Safety Features Extraction}
After estimating the depth and speed of the PV, \ac{TTC} and \ac{THW} are computed as follows:
\begin{equation}
  \mathrm{TTC} = \frac{|d|}{|v_{\mathrm{rel}}|}, 
  \quad 
  \mathrm{THW} = \frac{d}{v_{\mathrm{TV}}},
\end{equation}
where \(d\) is the relative distance between the TV and PV, \(v_{\mathrm{rel}} = v_{\mathrm{TV}} - v_{\mathrm{PV}}\) is their relative velocity, and \(v_{\mathrm{TV}}\) is the TV’s velocity. \ac{TTC} represents the time remaining before a potential collision if both vehicles maintain their current speeds, while \ac{THW} indicates the temporal gap between the two vehicles. These values are transmitted via TCP/IP to the communication module and subsequently relayed to the prediction module.

\subsection{Results}
\begin{figure}[ht]
  \centering
  \begin{subfigure}[t]{0.48\textwidth}
    \includegraphics[width=\textwidth]{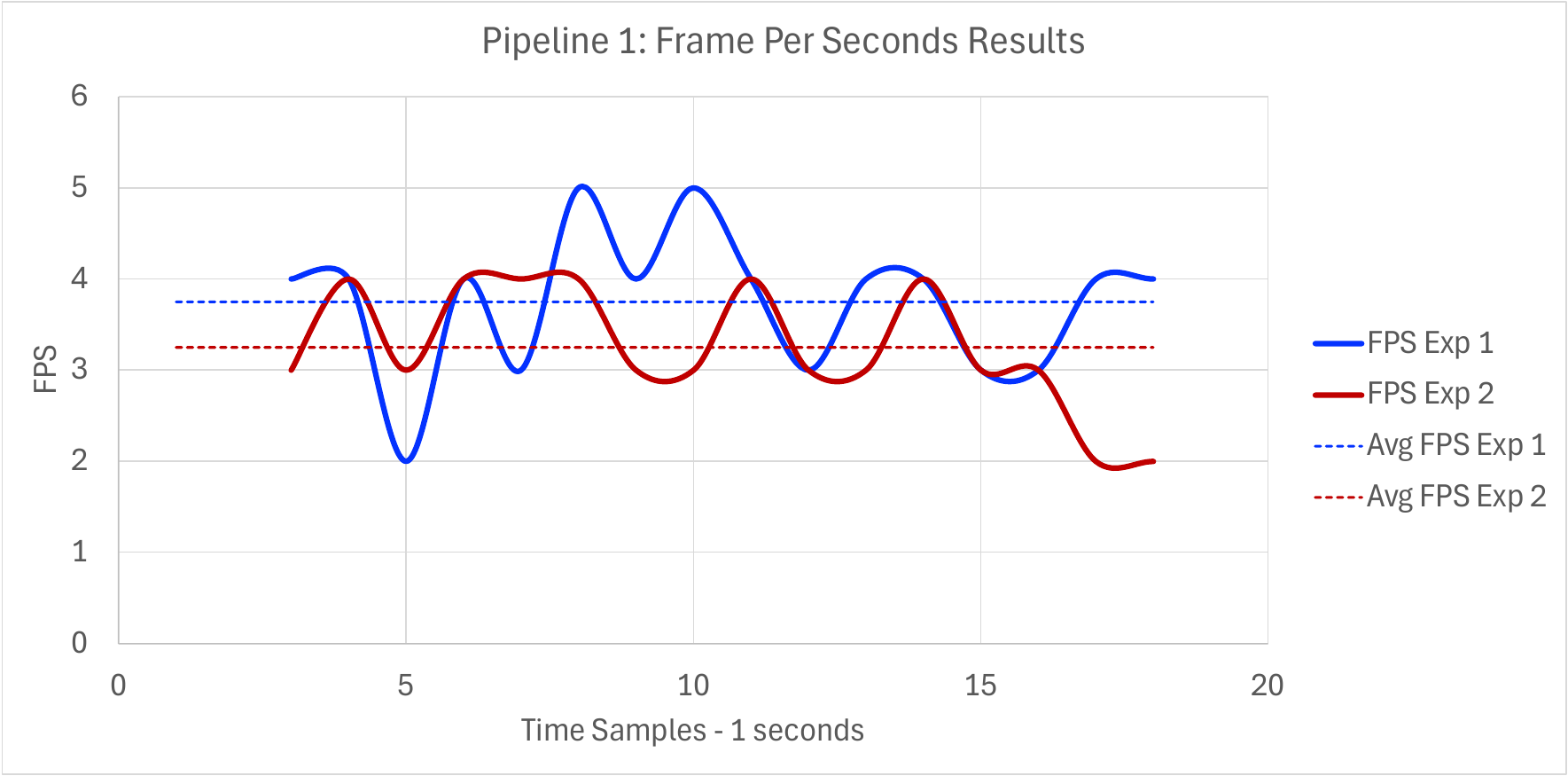}
    \caption{Pipeline 1}
    \label{fig:Pipeline_1_FPS}
  \end{subfigure}
  \hfill
  \begin{subfigure}[t]{0.48\textwidth}
    \includegraphics[width=\textwidth]{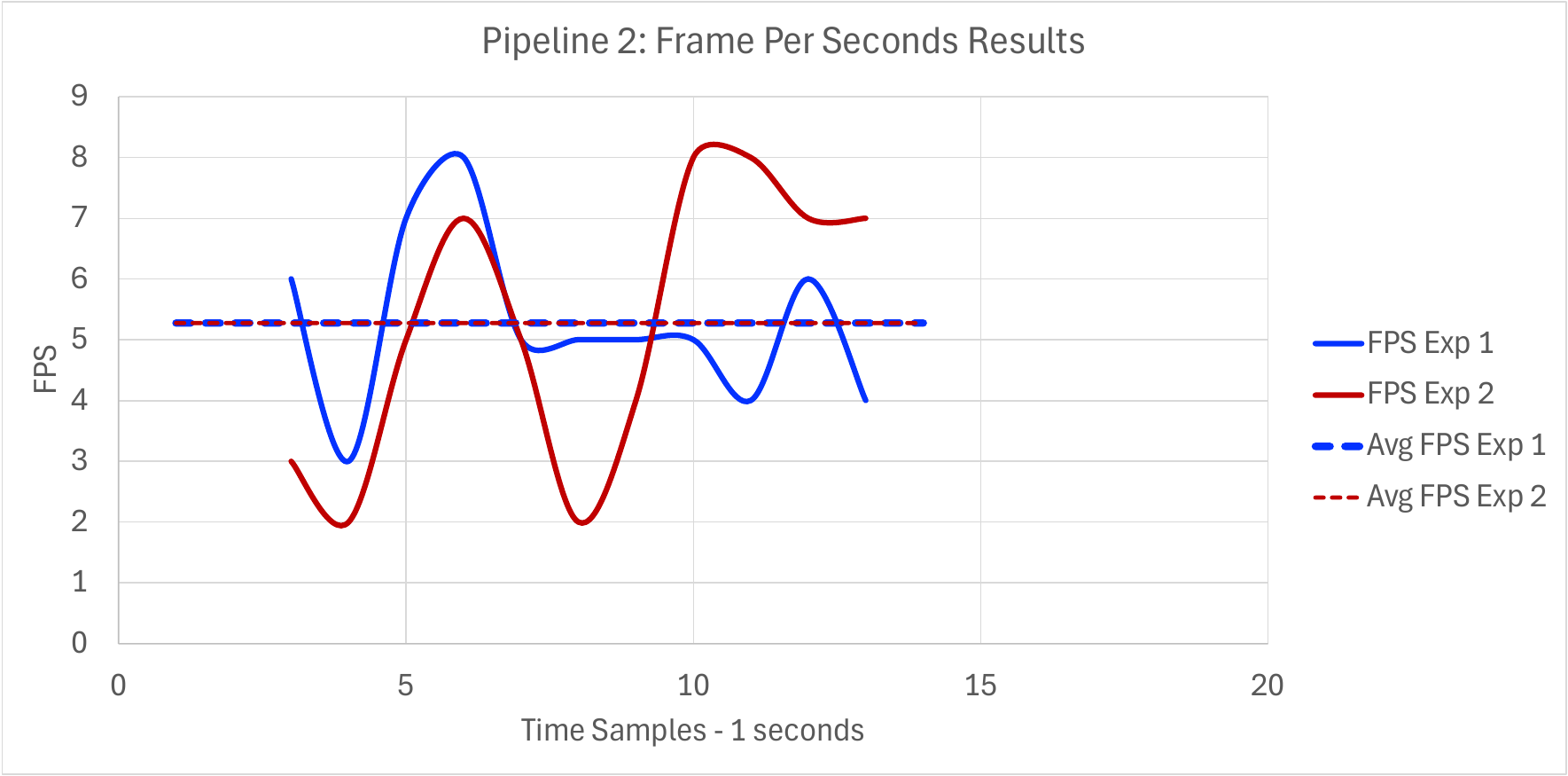}
    \caption{Pipeline 2}
    \label{fig:Pipeline_2_FPS}
  \end{subfigure}
  \caption{Comparison between the investigated pipelines Performance in terms of the \ac{FPS} during online Testing}
  \label{fig:Pipelines_FPS}
\end{figure}

\begin{figure}[ht]
  \centering
  \begin{subfigure}[t]{0.32\textwidth}
    \includegraphics[width=\textwidth]{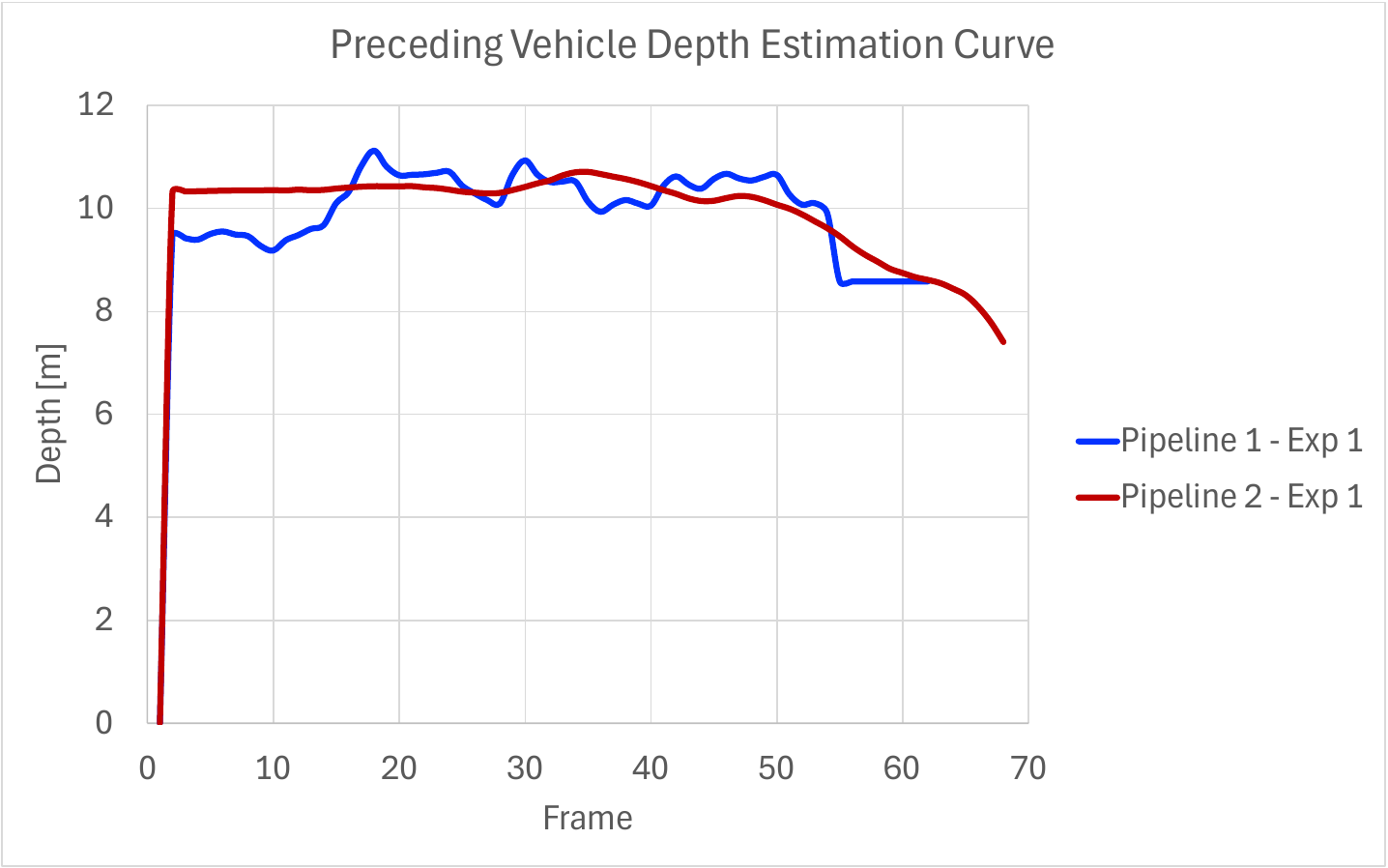}
    \caption{PV Depth}
    \label{fig:PV_Depth}
  \end{subfigure}
  \hfill
  \begin{subfigure}[t]{0.32\textwidth}
    \includegraphics[width=\textwidth]{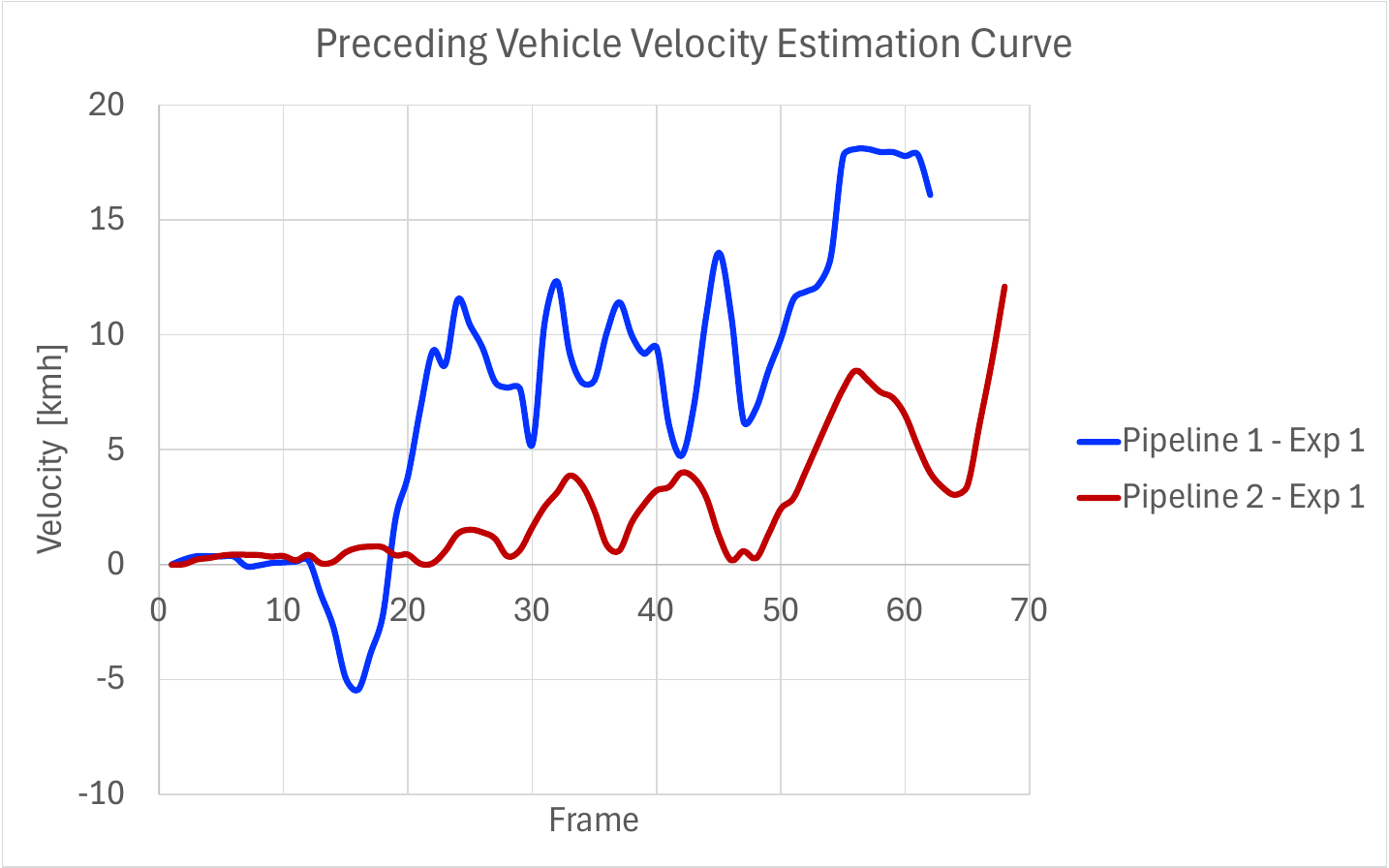}
    \caption{PV Velocity}
    \label{fig:PV_Velocity}
  \end{subfigure}
  \hfill
  \begin{subfigure}[t]{0.32\textwidth}
    \includegraphics[width=\textwidth]{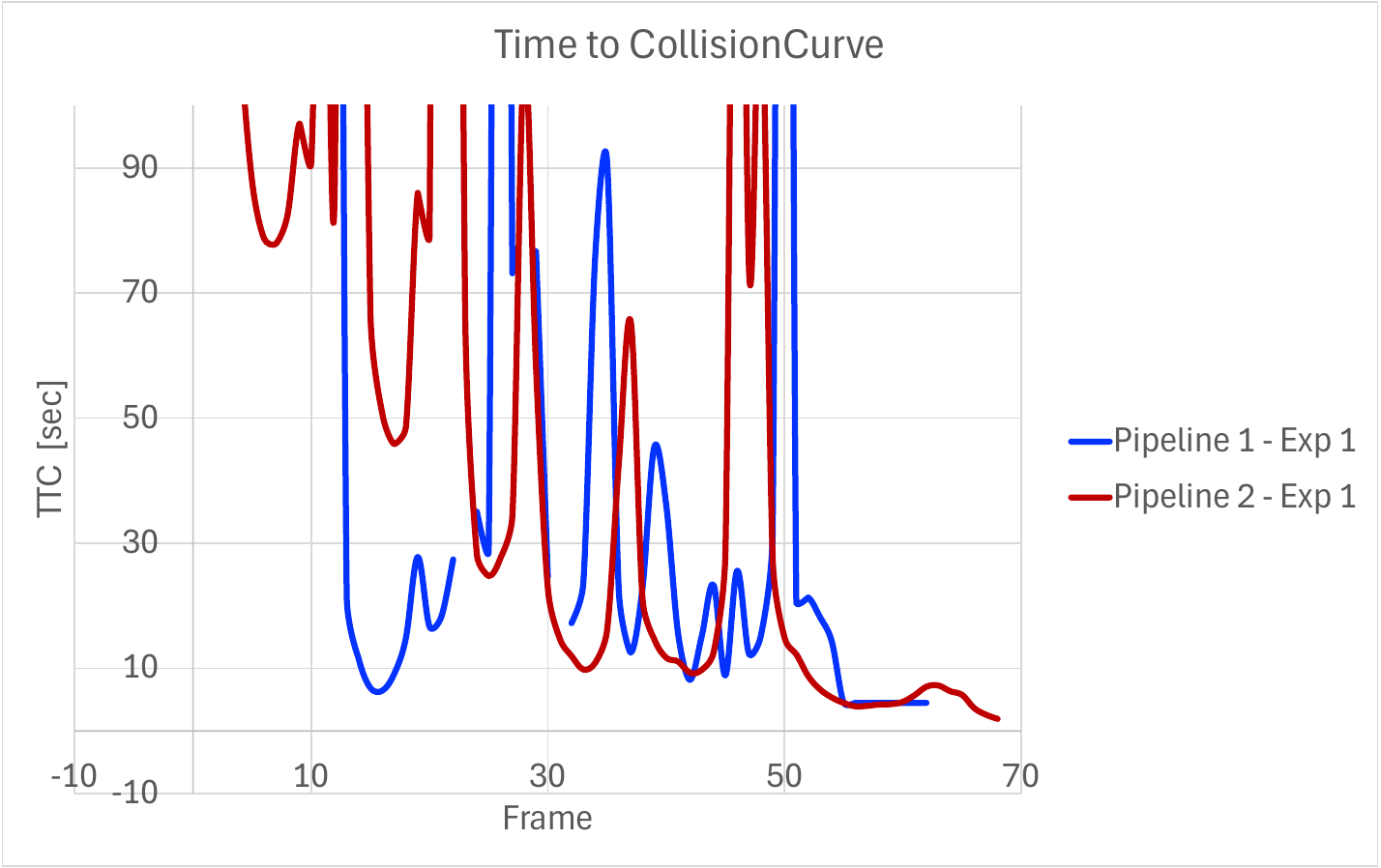}
    \caption{TTC}
    \label{fig:TTC}
  \end{subfigure}
  \caption{Perception Pipeline 1 and 2 results (a) The Depth estimation of the PV using RAFT vs ZED SDK, (b) The Velocity Estimation of the PV based on the estimated depth and relative velocity, and (c) The final estimated Time-to-Collision between the TV and PV.}
  \label{fig:Pipelines_Est}
\end{figure}

The performance of the perception module is assessed under both investigated pipelines. Both pipelines have been experimented twice in the full integrated system with the prediction module. Both experiments are basically under the same scenario. However, it is repeated to prove the repeatability and robustness of the overall system. It is important to note that even if the experiment is basically the same, it is impossible to yield the same exact results since 2 of the 3 vehicles are human-driven. That means, they will have different velocity and acceleration profile, hence slightly different results. In this section, we will mainly focus on illustrating the quantitative assessment of the perception module only. First in terms of the speed of the model, \Cref{fig:Pipelines_FPS} shows the variation in \ac{FPS} throughout the experiment, which lasted approximately $15$–$17$ seconds. It can be seen that pipeline 2 is running with higher \ac{FPS} with maximum of $8\ac{FPS}$ and average of  $5.3 \ac{FPS}$ compared to maximum of $5\ac{FPS}$ and average of $3.75 \ac{FPS}$ by pipeline 1. However, it is important to note that these results are recorded for the online testing of the experiments with the laptops onboard of the vehicle with no power charging. These average \ac{FPS} values significantly increase in the offline testing to reach $10-15 \ac{FPS}$ which was considered as a selection criteria for the used models. It is important to note that in both experiments for pipeline 2, the average \ac{FPS} value is consistent unlike the values obtained by pipeline 1. From this we can conclude that pipeline 2 is faster than pipeline 1 due to the lightweight models and the reduction of the detected object by using the ROI-Segmentation model.
To assess the estimators and the features extracted from both pipelines, two experiments with similar experimental setups were conducted. By comparing the estimated distances in experiment 1, it was found that the depth values from both pipelines were close to each other over the experiment duration. As shown in \Cref{fig:PV_Depth}, the vehicles maintained an approximate gap of $10 \text{m}$.

Moving to analyzing the estimated PV velocity using both trackers, \Cref{fig:PV_Velocity} shows that in pipeline 1 the vehicles moved with higher speeds than in pipeline 2. Still, in both cases the vehicles kept almost the same average gap of about $10,\text{m}$. This led to a riskier driving profile in pipeline 1, as confirmed by the lower \ac{TTC} values in \Cref{fig:TTC} compared to pipeline 2, where the lower speeds indicated less risk. It can also be seen from \Cref{fig:TTC} that the minimum \ac{TTC} occurred at the end of the experiment, which corresponds to the moment when the PV was braking and the TV was making the risky maneuver to avoid a collision.

\subsection{Findings}
\begin{itemize}
    \item \textbf{Computational Performance:}  
    \begin{itemize}
        \item RAFT-Stereo required significant processing power, achieving average $3 \ac{FPS}$ during the online experiment, while ROI-based segmentation was more efficient average $5.3 \ac{FPS}$. That confirms the importance of working with specified ROI.
        \item Offline runs confirmed that hardware load was the bottleneck, as both pipelines reached $10-15 \ac{FPS}$ with stable power.  
        \item This highlights the importance of hardware-aware pipeline design for scalable real-time deployment.  
    \end{itemize}

    \item \textbf{Depth Estimation Observations:}  
    \begin{itemize}
        \item The ZED 2 stereo camera delivered reliable depth within the $0.3-12m$ range, with high accuracy ($<1\%$ error at $3m$).  
        \item Beyond $12m$, measurements became increasingly noisy, reinforcing the need to prioritize mid-range detection in cooperative perception.  
    \end{itemize}

    \item \textbf{Environmental Insights:}  
    \begin{itemize}
        \item Lighting strongly influenced stereo matching: shaded or evenly lit scenes produced stable results, while direct sunlight and glare introduced distortions as shown in \Cref{fig:camError}. This will be considered as a challenge specifically for sunny countries to experiment during day light and of course it is always a challenge to experiment at night using vision-based perception. This only gave us the sunset time to be the most suitable time for experimentation.
        \begin{figure}[ht]
        \centering
        \includegraphics[width=0.75\textwidth]{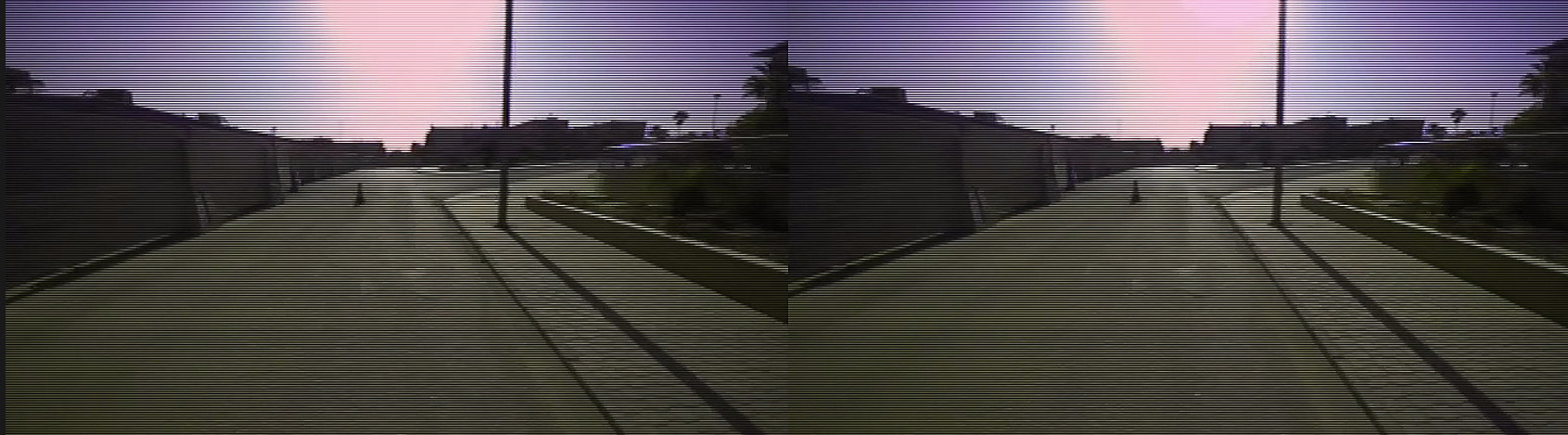}
        \caption{Distorted camera view.}
        \label{fig:camError}
        \end{figure}
        \item GPS readings were occasionally disturbed by electromagnetic interference like in the near high-voltage stations areas or narrow urban areas with high buildings, making throttle-based velocity estimation a more practical alternative in such contexts. Also, taking direct measurement is also a challenge since it is always affected by the battery charging level.  
    \end{itemize}

    \item \textbf{Camera and Sensor Behavior:}  
    \begin{itemize}
        \item ZED 2 performance was sensitive to focus drift and calibration issues in extreme lighting conditions.  
        \item Maintaining stable results required scenario-aware deployment to avoid scenacrios as reflective ground surfaces and testing during mild-light periods.  
    \end{itemize}

    \item \textbf{Pipeline-Level Insights:}  
    \begin{itemize}
        \item Pipeline 2 demonstrated more consistent \ac{FPS} and smoother perception under resource-constrained settings.  
        \item Pipeline 1 produced richer kinematic details but was more vulnerable to dropped frames and delays in sudden object motion.  
        \item Both pipelines successfully enabled computation of \ac{TTC} and \ac{THW}, supporting cooperative perception for safety-critical applications. However, they are also function in the accuracy of the estimations which encounter challenges already.
    \end{itemize}
\end{itemize}

\section{Communication Module}
Reliable communication between perception and prediction modules is essential for real-time cooperative driving. In this study, we evaluated two strategies: direct vehicle-to-vehicle (V2V) links for minimal latency, and a relay-based architecture for improved modularity and scalability.

\subsection{Direct V2V Communication}
The first strategy was based on direct Vehicle-to-Vehicle (V2V) communication, where the perception module, running on a laptop \textit{(Device 1)} in the TV that sends the extracted features directly to the prediction module hosted on either a Jetson Nano or a Raspberry Pi in the EV. In this configuration, only the communication delay was measured, while the prediction time was excluded. Although V2V communication allowed for direct exchange of perception information, it revealed several limitations. The perception and prediction modules were tightly coupled, such that any modification in the perception pipeline required corresponding adjustments in the prediction module. This limitation becomes even more critical if the concept is generalized to a larger number of vehicles: imagine a situation with many perception vehicles transmitting to many prediction devices. A single change in the perception pipeline would then propagate to all connected prediction modules, creating significant overhead and limiting flexibility. Furthermore, the direct wireless link was not always reliable. When the distance between vehicles increased, the Wi-Fi connection occasionally dropped, leading to communication interruptions and loss of data exchange.

\subsection{Relay-Based Communication}
To address these issues, a second strategy was implemented using a relay-based architecture, where a laptop \textit{(Device 2)} acted as an \ac{RSU}. In this case, the perception module \textit{(Device 1)} transmitted numerical features to the \ac{RSU} \textit{(Device 2)}, which handled preprocessing and conversion into linguistic categories before forwarding the reformatted data to the prediction device. This design decoupled perception from prediction, allowing each module to focus solely on its intended function: the perception module concentrated on sensing and feature extraction, the relay processed and formatted the data, and the prediction device executed inference. The modular structure ensured that changes in one module did not propagate to others, improving system maintainability. More importantly, this modularity scales better to multi-vehicle environments. Instead of requiring updates across a large number of prediction modules whenever the perception pipeline changes, only the relay would need to be updated, while all connected vehicles continue to receive standardized information. In this way, perception and prediction are fully decoupled, and modifications in one do not force changes in the other. Furthermore, placing a relay between vehicles stabilized the communication link by reducing the likelihood of dropped connections, since the relay acted as a fixed and reliable communication point. Finally, this approach offers a path toward scalability. In a multi-vehicle scenario, it would be impractical for a single perception module to establish and maintain direct links with multiple prediction devices. Instead, the relay can process the data once and distribute it to the appropriate vehicles. While our experiments involved only one perception and one prediction vehicle, we expect that this design would scale more naturally to larger traffic environments as shown in \cite{wang2021cooperative, ali2025enhancing}.

\subsection{Experimental Setup and Delay Measurement}
All communication was established over the TCP/IP protocol using the available Wi-Fi interfaces on each device. The Raspberry Pi 3B+ is equipped with an onboard Broadcom Wi-Fi chipset limited to 802.11n at 2.4 GHz. The Jetson Nano 2 GB employs an external USB Wi-Fi dongle supporting 802.11ac at 5 GHz, which enables faster and more stable links. The client device \textit{(Device 1)} is a laptop running Windows, while the server laptop \textit{(Device 2)} runs Linux; both laptops feature integrated dual-band adapters with 802.11ac/ax capability at 5 GHz. All communication in this work was established using the TCP/IP protocol. To quantify the communication delay, a Round-Trip Time (RTT) measurement procedure was employed. In this approach, a device sends a signal to the other device, which immediately echoed it back. The measured RTT in that case represents the total travel time of the message between the two devices in both directions, and the one-way communication time was obtained by dividing this value by two. This procedure was applied consistently across all tests. It was used for the direct V2V communication between the perception and prediction modules, as well as for the relay-based strategy, which includes both the communication from the perception laptop \textit{(Device 1)} to the relay \textit{(Device 2)} and the communication from the relay to the prediction module.

\subsection{Results}
\Cref{tab:relay} summarizes the measured communication overhead between the different device pairs in the experiment. The first column corresponds to the communication between the laptop client \textit{(Device 1)}, which serves as the perception module, and the Raspberry Pi. This link required 4.0 ms per iteration. The second column represents the communication between \textit{Device 1} and the Jetson Nano, which took 3.25 ms per iteration. Both of these cases fall under the direct V2V strategy, where the Jetson achieved a 0.75 ms per-iteration advantage compared to the Raspberry Pi. In the relay-based strategy, the process changes. The perception laptop \textit{(Device 1)} first communicates with the Linux laptop \textit{(Device 2)}, which acts as the relay (3.5 ms per iteration). After processing and formatting the data, the relay forwards it to the Jetson Nano (3.25 ms per iteration). The total communication overhead for this two-step process is 7.25 ms per iteration. Although this is nearly double the latency of the direct V2V case, the relay-based approach (as previously mentioned) provides important benefits in terms of modularity and scalability, making it more suitable for multi-vehicle environments.

\begin{table}[]
\caption{Measured communication overhead between device pairs. Direct V2V communication required 4.0 ms (\textit{Device 1}–Pi) and 3.25 ms (\textit{Device 1}–Jetson) per iteration, while the relay-based path (\textit{Device 1}–\textit{Device 2}–Jetson) increased the time to 7.25 ms. Although the relay nearly doubles the delay, it provides the advantages of modularity and scalability.}

\label{tab:relay}
\begin{tabular}{|c|c|c|c|}
\hline
Approach & \textit{Device 1} - Pi & \textit{Device 1} - Jetson & \textit{Device 1} - \textit{Device 2} \\ \hline
time ms/itr      & 4           & 3.25               & 3.5               \\ \hline
\end{tabular}
\end{table}
\subsection{Findings}
\begin{itemize}

    \item Direct V2V links achieved the lowest latency (3.25–4.0 ms), but tightly coupled perception and prediction, making the system inflexible to updates or changes. The real bottleneck is not milliseconds of delay but the system’s scalability and flexibility.  

    \item Wireless reliability in direct V2V was unstable; links dropped when the vehicle distance increased, making it unsuitable for consistent outdoor operation. Connection stability is therefore more critical than raw latency in real-world deployments.  

    \item The relay-based approach introduced slightly higher latency ($\sim$7.25 ms) but enabled full decoupling of perception and prediction, which is critical for modularity and maintainability.  
    
    \item Relay communication scales better to multi-vehicle environments: only the relay requires updates when perception changes, avoiding the rapid growth of direct links as the number of vehicles increases.

    \item Direct V2V links were unstable over distance, with frequent connection drops. The relay, acting as a fixed RSU, provided more robust and reliable communication. 

    \item For multi-vehicle environments, the relay design is preferable since only the relay must be updated when the perception pipeline changes, avoiding the impracticality of maintaining many direct links. 

    \item Device capabilities influenced communication performance: the Jetson Nano with 802.11ac (5 GHz) outperformed the Raspberry Pi with 802.11n (2.4 GHz), showing that communication quality is shaped not only by architecture but also by Wi-Fi hardware and frequency bands.  

    \item Despite nearly doubling the delay, the relay kept communication overhead within practical real-time limits. The extra relay overhead is negligible in practice given the safety margins gained from modularity and stability.  

    \item In conclusion, the relay-based approach offers the most practical balance of scalability, robustness, and modularity, making it better suited than direct V2V for cooperative perception in real-world deployments.

\end{itemize}

\section{Prediction Module}
\subsection{KGE Training}
A knowledge graph is a structured representation of information where entities are expressed as nodes and their semantic relations are expressed as edges, typically in the form of triples of the type ⟨subject, relation, object⟩. In the context of vehicle intention prediction, such a representation allows numerical data describing vehicle motion to be translated into semantic facts, such as “the vehicle has high-risk \ac{TTC} with the preceding car” or “the lateral acceleration is close to zero.” Unlike raw numerical values, these semantic descriptors can be directly interpreted by humans and can serve as the foundation for reasoning systems. In order to use these graphs for prediction tasks, knowledge graph embeddings are introduced. Knowledge graph embedding models project the entities and relations of the graph into a continuous low-dimensional vector space while preserving semantic proximity. Entities and relations that are strongly associated in the data are placed closer together in the embedding space, whereas unrelated elements are placed further apart. This transformation makes it possible to apply machine learning and probabilistic reasoning on top of symbolic knowledge.

The prediction pipeline builds on twelve input features that describe the kinematic and contextual state of the scene. These include the lateral velocity and acceleration of the TV; \ac{TTC} with the preceding, left-preceding, right-preceding, left-following, and right-following vehicles; lane identifier; position within the lane; \ac{THW} with the PV; the lane with the highest frontal gap; and the lane with the highest attraction score. Because these variables are raw numerical quantities and not inherently interpretable, they are first converted into linguistic categories using thresholding criteria. The conversion is carried out according to two principles. For features such as lateral velocity, the distribution of the data was analyzed, the observed values followed a normal distribution, and by applying the mean and standard deviation, the values were divided into three categories: \emph{moving left}, \emph{moving straight}, and \emph{moving right}. Other features, such as \ac{TTC}, were categorized into \emph{high risk}, \emph{medium risk}, and \emph{low risk} according to thresholds reported in the literature. After this conversion, the categories are reified into triples (<subject, relation, object>) and stored in a tabular form in a Comma-Separated Values (CSV) file. Each row of the file corresponds to a fact that describes the scene. For example: "vehicle, LATERAL\_ VELOCITY\_ IS, movingLeft", "vehicle, TTC\_ WITH\_ PRECEDING\_ VEHICLE\_ IS, highRisk" , and "vehicle, INTENTION\_ IS, leftLaneChange". These triples are constructed according to a formal ontology that defines all entities and possible relations. Further details on the knowledge graph construction and complete description of the ontology and its instances are available in \cite{manzour2025explainable}. For embedding, the TransE model was used via the AmpliGraph library. The embedding size was set to $100$, the optimizer was Adam with a learning rate of $0.0005$, and training used a batch size of $10,000$. A self-adversarial loss function was applied with negative sampling at a $5:1$ ratio, and early stopping monitored using the Mean Reciprocal Rank (MRR).

\subsection{Bayesian Inference}
Once the graph is constructed and embedded, Bayesian inference is carried out on top of the learned embeddings to perform the prediction. The task is to compute the probability of each possible maneuver hypothesis H given the observed evidence E formed by the twelve linguistic inputs. This follows Bayes’ theorem:
\begin{equation}
P(H \mid E) = \frac{P(H) \cdot P(E \mid H)}{P(E)}
\end{equation}
Here, $P(H \mid E)$ is the posterior probability of a maneuver given the evidence. $P(H)$ is the prior probability of the maneuver, reflecting its baseline likelihood before considering current evidence, and is informed by statistical regularities captured in the embeddings. $P(E \mid H)$ is the likelihood, i.e., the probability of observing the evidence if the maneuver were true. Finally, $P(E)$ normalizes the result so that the posteriors over all hypotheses sum to one.

An example that illustrates the reasoning can be like the following. Suppose the system considers three hypotheses: lane keeping, left lane change, and right lane change. Initially, the prior favors lane keeping, since it is generally more common. If the \ac{TTC} with the PV is very small (high risk), the likelihood of lane keeping decreases, while the likelihood of either lane change increases. The posterior $P(H \mid E)$ then shifts accordingly. Next, if the \ac{TTC} with the right-following vehicle is also high risk, while that with the left-following vehicle is low risk, the evidence weakens the probability of a right lane change and strengthens that of a left lane change. At this point, the posterior favors the left lane change hypothesis. Through this sequential updating process, all available inputs (lateral motion,\ac{TTC} values, \ac{THW}, lane gaps, and attraction scores) contribute to refining the probability of each maneuver. The final posterior distribution represents the combined effect of all evidence, and the maneuver with the highest posterior probability is selected as the predicted intention of the TV. The strength of this approach lies not only in its predictive ability but also in its interpretability. Because each step of the inference process is explicit, one can always trace back how the evidence led to the final decision: the prediction of a left lane change can be justified by pointing to the high-risk \ac{TTC} with the PV, which reduced the probability of lane keeping, and to the low-risk conditions on the left, which increased the probability of a leftward maneuver. In this way, the Bayesian framework ensures that predictions are both accurate and transparent, qualities that are essential for safety-critical applications in autonomous driving.

\subsection{Results}
The model was trained on both the HighD dataset (safe lane changes) and the CRASH dataset (risky lane changes). It achieved an f1-score of 95\% and 90.0\% for safe maneuvers with two and four-second anticipation horizon respectively, and 95\% and 91.5\% for risky maneuvers under the same horizons. When both safe and risky maneuvers were combined in a mixed setting, the model reached an f1-score of 95\% and 89.4\%, showing consistent performance across scenarios. Compared with prior studies, our results are competitive with state of the art deep learning models that predict only safe lane changes. For example, transformer-based approaches \cite{gao2023dual} on HighD reached 98–95\% f1-scores, and XGBoost model used by \cite{} on HighD reached 99–92\% f1-scores. Both models predicted short horizons (0.5–2s), lacked interpretablity and transparency, and did not consider risky situations and didn't extend their horizon to 4 seconds. Our model, while slightly lower at very short horizons, maintains high accuracy up to four seconds, provides traceable Bayesian reasoning, and uniquely addresses risky near-crash lane changes. These results demonstrate that the approach not only performs well in standard safe scenarios but also extends predictive capability to high-risk conditions, making it more practical for safety-critical driving applications.

\subsection{Hardware Validation}
In terms of hardware validation, the model was evaluated on three different computing devices: a Lenovo laptop equipped with an NVIDIA RTX 4090 \ac{GPU} (16 GB VRAM, 32 GB RAM, and a 1 TB SSD), a Raspberry Pi 3 Model B+, and an NVIDIA Jetson Nano with 2 GB of memory. On the laptop, the model was tested in an online setting and achieved a speed of approximately three to four frames per second. However, it was not feasible to run the online version of the model directly on the Raspberry Pi or the Jetson Nano due to their more limited computational capacity.

To overcome this limitation, we exploited the fact that the model operates on linguistic variables. Instead of executing the full model online, we precomputed all feasible combinations of the twelve input features and their corresponding predictions. For example, the feature “\ac{TTC} with the PV” can be classified into three categories: low risk, medium risk, or high risk. Considering similar categories for all twelve features, a large set of possible input combinations can be generated. Some of these combinations, however, are not physically feasible. For instance, if the TV is already in the leftmost lane, it is not possible to assign the category “left lane has the highest frontal gap,” because no left lane exists. After removing all such infeasible cases, approximately 212,000 valid combinations remained. These combinations were first stored in a CSV file and loaded into a Pandas DataFrame. In this setup, when a prediction is required, the system searches by filtering row by row: each of the twelve inputs is matched against its corresponding column (for example, \textit{Input1 == x, Input2 == y, … Input12 == z)} until the single row that satisfies all twelve conditions is found, and the corresponding prediction is returned. This type of search behaves like an \textit{O(n)} operation, where \textit{n} is the number of rows in the file, because as the CSV grows larger, more records must be checked before the match is found. This is why prediction with CSV search slows down as the number of stored combinations increases. The first row in \Cref{tab:computational_time} reports the computational time, in seconds per iteration, of the three devices when using this CSV file approach. As expected, the laptop achieved the lowest computational time (0.075 seconds per iteration) due to its superior hardware, but this entry is mainly included for comparison. Since the final system must run on an embedded platform, the relevant comparison is between the Raspberry Pi and the Jetson Nano. From the results, it is clear that the Jetson Nano provides better computational performance than the Pi when using the CSV-based search, Jetson outperformed the Pi by about 31\%.
To improve efficiency, the combinations were also stored in a lookup table, implemented as a dictionary structure. In this representation, all twelve inputs are concatenated into a single string key at the time of storage. At inference time, the twelve current linguistic inputs are concatenated in the same way, and the dictionary is queried directly using this key to return the prediction. This avoids the need for sequential filtering across twelve separate columns, replacing it with a single direct lookup.From a computational standpoint, the dictionary approach has constant-time complexity \textit{O(1)}, meaning that the search time does not increase with the number of stored combinations. The results of the lookup approach, shown in \Cref{tab:computational_time}, indicate much better computational time across all devices. The laptop reduced its computational time from 0.075 seconds to $1.4e^{-7}$ seconds, corresponding to a speed-up of $(10^5\times)$. The Raspberry Pi dropped from 1.16 seconds to $9.4e^{-6}$ seconds, and the Jetson Nano from 0.80 seconds to $2.4e^{-6}$ seconds.
This means the lookup approach provided a speed-up of nearly $(10^5\times)$ on the Pi and over $(3*10^5\times)$ on the Jetson, confirming that lookup tables make real-time deployment feasible. Moreover, the Jetson consistently outperforms the Pi, making it the preferred option for in-vehicle integration. To further analyze scalability, the experiment was repeated with a reduced dataset representing only two-lane roads, resulting in approximately 18,000 feasible combinations instead of 212,000. In this smaller configuration, the CSV search became faster than in the previous case as shown in the 3rd row of \Cref{tab:computational_time}, as expected due to the reduced number of rows to filter. However, the lookup table continued to outperform the CSV approach and, as anticipated, maintained nearly identical performance to the larger configuration as shown in the 4th row of the table. This confirms that lookup-based prediction provides stable speed, while CSV search depends heavily on dataset size.

\begin{table}[ht]
\centering
\caption{Computational time (seconds per iteration) for CSV and lookup search across devices. Lookup tables consistently achieve much lower latency than CSV search, and maintained performance regardless of dataset size.}
\label{tab:computational_time}
\begin{tabular}{|c|c|c|c|c|}
\hline
\begin{tabular}[c]{@{}c@{}}Number of\\ Combinations\end{tabular} & Approach & Laptop & Pi & Jetson \\ \hline
\multirow{2}{*}{212K}                                           & CSV      & $0.075$      & $1.16$  & $0.80$      \\ \cline{2-5} 
                                                                & LOOKUP   & $1.4e^{-7}$      & $9.4e^{-6}$  & $2.4e^{-6}$      \\ \hline\hline
\multirow{2}{*}{18K}                                            & CSV      & 0.0067      & 0.11  & 0.070      \\ \cline{2-5} 
                                                                & LOOKUP   & $1.4e^{-7}$      & $9.4e^{-6}$  & $2.4e^{-6}$     \\ \hline
\end{tabular}
\end{table}

\subsection{Findings}
\begin{itemize}

    \item  Unlike many black-box deep learning approaches, the use of linguistic information along with Bayesian reasoning ensures interpretability and transparency, making it suitable for safety-critical driving applications.

    \item The laptop offered the highest efficiency (fastest execution), but this serves only as a comparison baseline and is not intended for real in-vehicle deployment.  
    
    \item Among embedded devices, the Jetson Nano consistently outperformed the Raspberry Pi (by $\approx$31\% under CSV search and by orders of magnitude under lookup table), making it the preferred option for onboard integration. 
    
    \item The CSV-based search was computationally heavy (up to 1.16 s/iteration on the Pi). By contrast, lookup tables reduced latency by $\approx 10^{5}\times$ to microseconds, achieved constant-time performance ($O(1)$), and maintained stable inference regardless of dataset size (tested at $\sim$212k vs. $\sim$18k combinations). This enables real-time and scalable cooperative prediction. 
    
    \item With prediction active, the EV anticipated the TV’s maneuver $\sim$4 s before the lane change, yielding smoothly and avoiding abrupt braking. Without prediction, the TV was forced into sharp braking, confirming that predictive reasoning directly improves safety and comfort in cooperative driving.   

    \item In conclusion, the Jetson Nano with lookup tables offers the best trade-off for real-time, interpretable prediction on embedded systems, directly supporting safer and smoother cooperative lane-change interactions.

\end{itemize}

\section{Planning and Control Module}
This section addresses the implementation details related to the control of the EV. In the upcoming subsections we would discuss the motion planning details, afterwards the sensors used in the EV setup will be discussed and finally the control implementation details are presented.
\subsection{Motion Planning}
Vehicular motion planning is required to be done for both longitudinal (velocity) and lateral (steering) motions. In this study, the scope of the EV motion considered requires it to be in a specific direction in its lane, without the need to shift lane. Hence, for a state-feedback lateral control was implemented maintaining the vehicle's heading constant in the same direction. To that end, constant feedback of the heading of the vehicle is provided, based on which minor incremental steering angles could be applied (when needed) to maintain the heading constant.\\
However, in this study, the longitudinal motion of the EV is necessary to be controlled in an online manner, which is achieved via manipulating the speed profile of the vehicle. The EV used in this study is a scaled 1:4 vehicle following the body dimensions of a BMW x3 vehicle. The vehicle achieves a maximum speed of 1.5m/s (around 5.4 km/h). The speed of the vehicle is controlled via manipulating the Pulse-Width-Modulation (PWM) method available on the low-level controller onboard, from 0-100\%. For the purpose of the experiments in this study, it is required to control the motion of the vehicle based on one of 3 states. In each time step, the vehicle receives the desired state from the prediction module, and updates the desired PWM accordingly. Firstly, the (accelerate) state, for which the vehicle's velocity PWM is increased incrementally by 4\% each sampling time (100ms). Secondly, the state (decelerate), in which the vehicle is required to slow down gradually. In this case the deceleration is done by decrementing the PWM values by 8\% each time step. This is done in order to increase safety. Finally, the (stop) state aiming to terminate the vehicle's speed instantaneously in case of emergency, in which the vehicle decelerates by its maximum possible value by setting the desired PWM value to 0\%.
\subsection{Onboard Sensors}
To be able to control the vehicle's motion to follow the desired motion planned, a set of onboard sensors were used. For the longitudinal motion, velocity feedback is obtained from a single optical encoder coupled with the vehicle’s rear wheel axis. This 600-pulse incremental rotary encoder, operating at 5V, provides feedback to a low-level onboard processor dedicated to speed control. From the pulses generated by this encoder, both the vehicle’s speed and traveled distance can be calculated and accordingly controlled. For the lateral motion, an onboard \ac{IMU} sensor version MPU6050 is used providing feedback only for the yaw angle rate. If the value of the yaw angle increased beyond a predefined margin of $\pm 2$ degrees from its original value at the beginning of the experiment, the steering wheel motor (servo motor) is activated to compensate this deviation. It is worth to mention that effect of the vibration from the road on the deviation of the vehicle is amplified for scaled vehicles than real-sized ones, and thus proper control of heading in case of other experiments that include lateral motion is a must. The readings of the IMU is provided to another low-level processor dedicated for the heading control.
\subsection{EV Control}
As mentioned in the previous subsections, lateral control in this study is done using a simple state-feedback control law aiming to maintain the heading of the vehicle constant. On the contrary, in this experiment, longitudinal control is essential to make sure that the vehicle is following the desired velocity profile. This is achieved by using a tuned \ac{PID} control for speed control of the vehicle based on the following set of equations. The velocity error in each iteration is calculated using the followig equation:
\begin{equation}
    e(k)=v_d(k)-v_a(k)
\end{equation}
where $e(k)$ is the velocity error in the current iteration, while the $v_d(k)$ and $v_a(k)$ are the desired and the actual velocity of the EV in the current iteration respectively. Accordingly, the proper action from the \ac{PID} controller is calculated using the following equation:
\begin{equation}
    x(k) = k_pe(k)+k_i\sum_{i=0}^{k} \frac{e(k)+e(k-1)}{2}T_s+k_d\frac{e(k)-e(k-1)}{T_s}
\end{equation}
where $x(k)$ is the calculated \ac{PID} output (before mapping), $k_p$, $k_i$ and $k_d$ are the proportional, integral and derivate \ac{PID} gains that tune the effect of each error component. $e(k)$ and $e(k-1)$ present the velocity error in the current iteration and the previous iteration respectively, and finally $T_s$ present the control loop sampling time (100 ms) in this study. After the \ac{PID} control action is calculated, proper mapping is required for it to convert it to the PWM acceptable range of value [0-255] and then sent to the speed controller. The desired speed is defined from the motion planning module based on the previously explained states (accelerate, decelerate or stop).
\subsection{Observations and Findings}
From the experiments conducted, the following observations could be extracted for the scaled EV:
\begin{itemize}
    \item Lateral control performance was completely different between in lab experiments and the real road experiment, this was attributed to the vibration from the asphalt road which as propagated heavily to the IMU sensor and altered its readings drastically. Accordingly, proper damping for the IMU sensor is required for outdoor experiments.
    \item Localization sensors commonly used for outdoor experiments were not suitable for the EV localization. Sensors such as GPS weren't able to accurately localize the scaled vehicle as the velocity range and distance traveled were small. Accordingly, utilization of other localization methods/sensors was needed. In this study we relied upon encoders however, it is possible also to rely on visual odometry (future work).
    \item During testing, the communication channel between the EV and other vehicles/devices was lost specially when the distance between them increased. At which case, the EV drifted during its motion, as it was not able to identify the proper course of action to follow. This was also observed for communication between onboard devices. These observations are highly impacted by many factors such as the presence of nearby devices, the distance from the RSU router, the charging level of the batteries supplying the devices, and others. To that end, having a fault-tolerant system for such modules is essential for better implementation of the demonstrated approaches specially in a Cooperative driving scenario.
    \item Another interesting observation was the crashing of some onboard processing units suddenly during the experiment when conducted during day time. After lots of investigation, it was attributed to the effect of high temperature. The thermal effect on the vehicles' sensors and processors have a huge effect on the stability of the systems, this has a huge impact on the future of implementing the approaches on vehicles operating in high temperature countries, such as the GUlf or Middle East and North Africa (MENA) region.
    \item Despite being a scaled vehicle, the EV successfully participated in an experiment with real-sized vehicles. Thus the scaled vehicles utilization provide a suitable platform that can be extended to demonstrate different cooperative vehicles behaviors interacting together in the future.
\end{itemize}

\section{System Integration}

\begin{figure}[ht]
\centering
\includegraphics[width=0.65\linewidth]{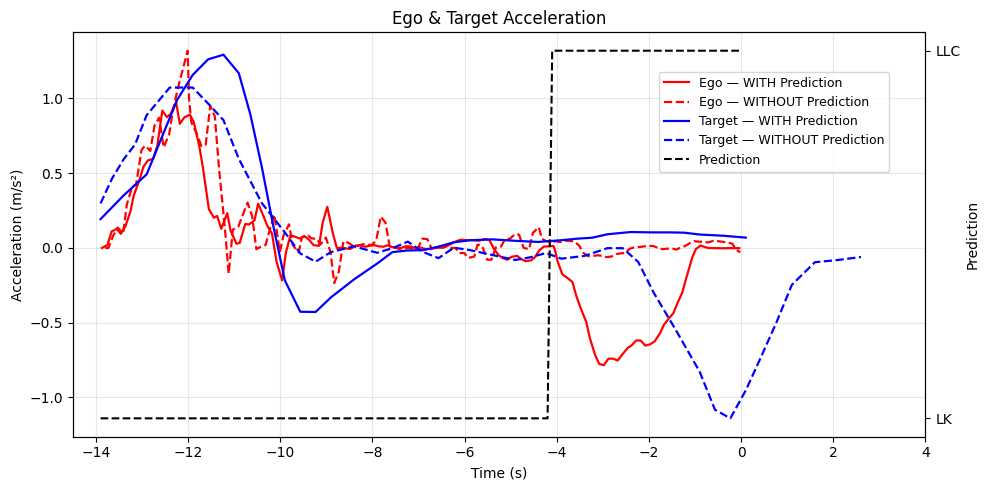}
\caption{Acceleration profiles of the EV (red) and the TV (blue) with and without the prediction model. The black dashed line indicates the maneuver prediction, and the gray vertical dashed line denotes the crossing moment at $t=0$.}
\label{fig:acceleration}
\end{figure}

\begin{figure}[ht]
\centering
\includegraphics[width=0.65\linewidth]{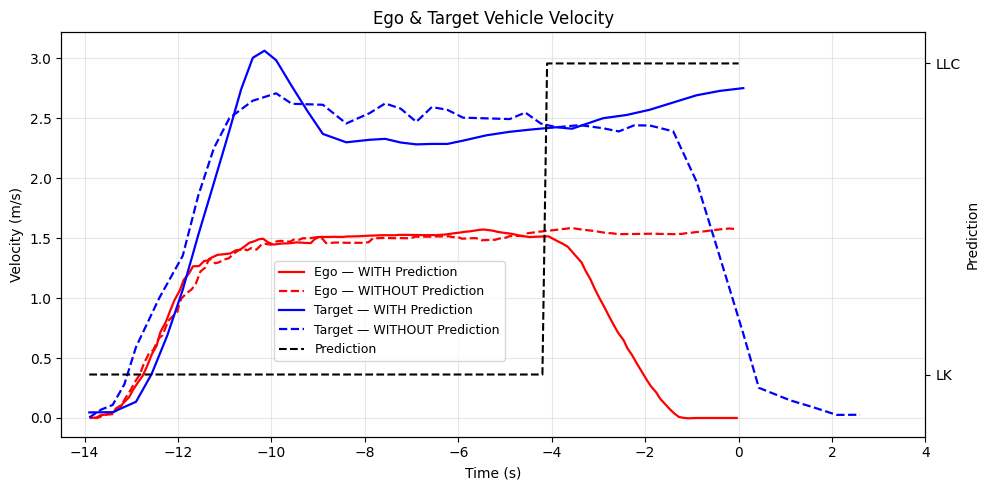}
\caption{Velocity profiles of the EV (red) and the TV (blue) with and without the prediction model. The black dashed line indicates the maneuver prediction, and the gray vertical dashed line denotes the crossing moment at $t=0$.}
\label{fig:velocity}
\end{figure}

After integrating all the modules, the two scenarios discussed in \Cref{sec:experimrntal_setup} were tested. In the first scenario, the prediction module was not integrated, while in the second scenario the prediction module was active in the system. To demonstrate the behavior of the integrated system, experimental trials were recorded and are available from multiple viewpoints. \Cref{tab:media} provides links to the two scenarios (with and without prediction), captured from different camera perspectives. These videos clearly show the different interaction outcomes depending on whether the prediction module is active. More views can be accessed from the following playlist link: \url{https://www.youtube.com/playlist?list=PLAeK3AuwxenFqIeAnKa9BLB8bDqEk8dNH}

The acceleration and velocity of both the EV and the TV were also recorded for these trials, as shown in \Cref{fig:acceleration} and \Cref{fig:velocity}. For consistency, $t=0$ corresponds to the moment when the TV crosses the lane marking, negative values indicate moments before the lane change, and positive values indicate moments after. When the prediction module was not integrated (dashed curves), the EV failed to anticipate the left-lane change maneuver of the TV. The EV continued straight ahead, which is reflected in the nearly constant acceleration of the red dashed ego-vehicle curve. This behavior resulted in a very small gap between the two vehicles, which did not permit the target to complete the lane change. Consequently, the TV was forced to brake sharply, visible as the sudden drop in the blue dashed target-vehicle curve close to the crossing moment. Such abrupt braking introduces potential safety risks (as the TV was about to collide with the PV) and discomfort for passengers. In contrast, when the prediction module was integrated (solid red curve for the EV and solid blue curve for the TV), the system anticipated the target’s left-lane change about four seconds before the TV reached the lane marking. The EV then began to yield smoothly, visible in the gradual decrease of its acceleration and velocity curves. As the EV comes to a stop, its acceleration gradually returns to zero, confirming a controlled and safe maneuver. This behavior created enough space for the TV to merge without braking. Accordingly, the TV’s acceleration and velocity remained nearly constant, showing no sudden deceleration.
\FloatBarrier
\begin{table}[!htbp]
\caption{Video links showing hardware validation results, with and without integrating the prediction module.}
\begin{center}
\begin{tabular}{|c|c|}
\hline
      Scenario View          & Link\\
\hline
WITH Prediction (View 1)   & \url{https://youtu.be/p4LKoaUsZpc}\\
\hline
WITHOUT Prediction (View 1)   & \url{https://youtu.be/BFVqhoT5Lck}\\
\hline
WITH Prediction (View 2)   & \url{https://youtu.be/8Dc6Agxgn8M}\\
\hline
WITHOUT Prediction (View 2)   & \url{https://youtu.be/U7PYoEeVnqY}\\

\hline
\end{tabular}
\label{tab:media}
\end{center}
\end{table}
The velocity profiles confirm this behavior. Without prediction (dashed curves), the EV maintained a nearly flat velocity profile, while the TV’s velocity dropped abruptly when it could not merge. With prediction (solid curves), the EV smoothly reduced its velocity to zero, as seen in the steady downward slope of the solid red curve, while the TV maintained its velocity without a sharp decrease, indicating that it could merge safely.

\section{Discussion of Findings and Conclusion}
This study examined a cooperative perception–prediction architecture for lane-change anticipation through real hardware deployment. The system was studied in real-world conditions with heterogeneous vehicles and embedded devices. By integrating sensing, communication, prediction, and control across different vehicles, the experiments demonstrated how cooperative information exchange can change the outcome of risky interactions. The experiments demonstrated that prediction changes the interaction outcome: when active, the ego vehicle smoothly yielded and enabled a safe merge, whereas without prediction the target vehicle was forced into abrupt braking. At the system level, several deep insights and bottlenecks emerged:
\begin{itemize}
    \item \textbf{Hardware Constraints:} Perception pipelines such as RAFT-stereo provided detailed kinematics but were limited by low \ac{FPS}. This highlights the trade-off between model richness and real-time feasibility.

    \item \textbf{Communication Reliability:} Direct V2V links offered the lowest delay but were unreliable over distance. The relay-based design, while adding ~3.5 ms latency, proved essential for modularity, scalability, and robust operation in multi-vehicle settings.

    \item \textbf{Prediction Scalability:} Lookup tables transformed inference into constant-time complexity, making real-time deployment feasible even on resource-limited devices. This design insight is crucial for scaling cooperative architectures.

    \item \textbf{Environmental Factors:} Lighting variability, GPS interference, and thermal effects on embedded boards were persistent issues, showing that perception and computation pipelines must be tailored to deployment environments.

    \item \textbf{System Integration:} The cooperative prediction fundamentally changed the traffic interaction outcome. Anticipation at a $\sim$4 s horizon enabled smoother and safer behavior, providing a level of foresight not achievable with reactive control alone.

\end{itemize}

In conclusion, the cooperative prediction pipeline was shown to be both functional and impactful in real-world conditions. However, the most significant bottleneck remains at the interface between perception and prediction. The reliance on computationally intensive perception models and the challenges posed by communication delays continue to affect the system’s consistency in real-time operation. To move toward broader deployment, future efforts should focus on optimizing perception for embedded hardware, improving the robustness of data transmission, and ensuring stability under varying environmental conditions. Even so, the experiments clearly show that cooperative prediction can make a decisive difference: even a few seconds of advance awareness is enough to change the course of interaction, enabling safer and more adaptive vehicle behavior.


\bibliographystyle{cas-model2-names}

\bibliography{manzour-et-al-references}


\newpage

\bio{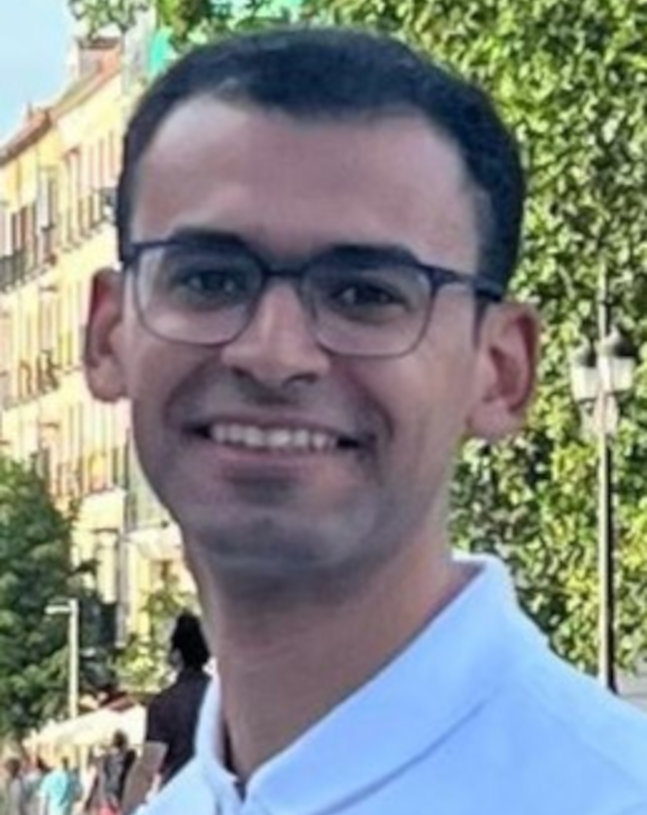}
Mohamed Manzour Hussien obtained his bachelor's degree in mechatronics from the German University in Cairo (GUC) in 2019. During his undergraduate studies, he had the opportunity to conduct his bachelor's thesis in the field of machine learning at the IFS (Institut für Schienenfahrzeuge) institute in Germany. After that, he worked as a lecturer assistant in the GUC till 2022. During this period, he completed his master's degree in the field of Intelligent Transportation Systems (ITS) at the Multi-Robot Systems (MRS) research group in the GUC in 2022. He focused on pedestrian behavior prediction. In 2023, he started his Ph.D. journey in the field of ITS at the INVETT (INtelligent VEhicles and Traffic Technologies) research group, University of Alcala, Spain. He is focusing on vehicle behavior prediction. In 2023, he won the IEEE ITSS student competition in the track of Driver Decision Prediction.
\endbio

\bio{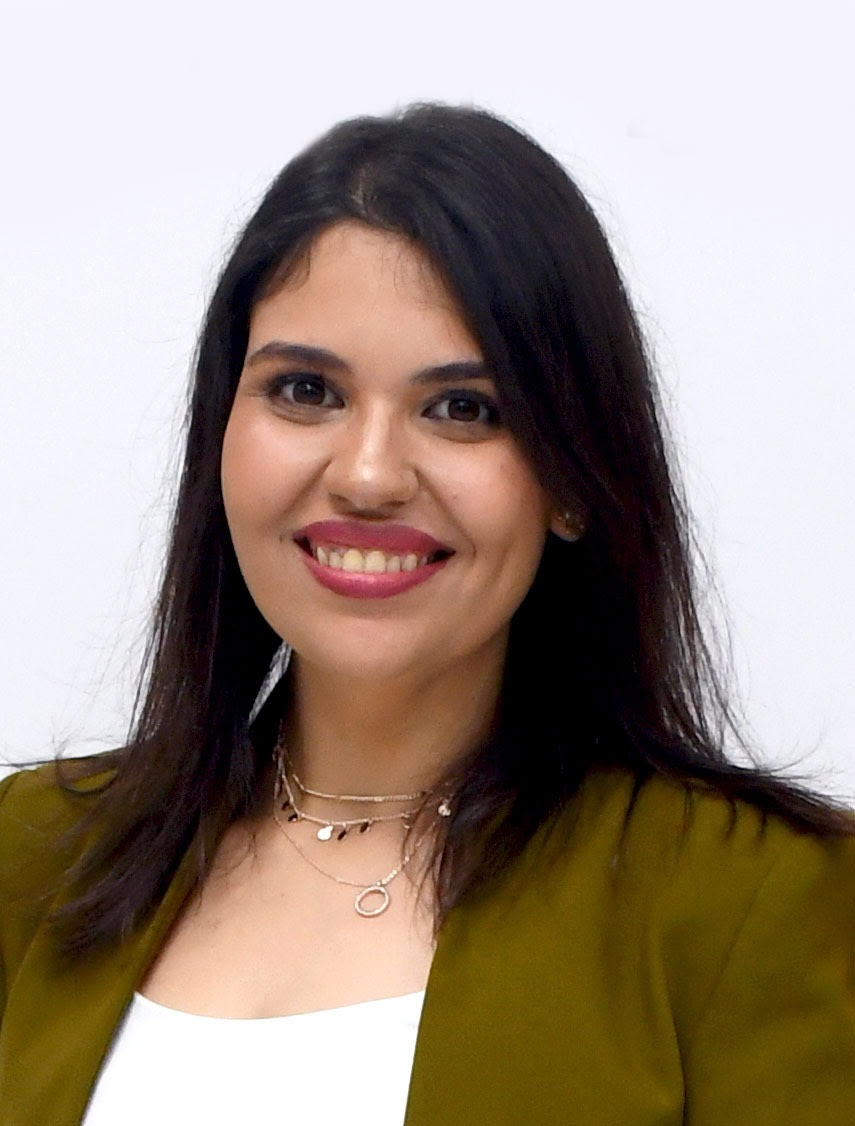}
Catherine M. Elias is a lecturer in the Computer Science and Engineering Department, Faculty of Media Engineering and Technology (MET), German University in Cairo (GUC). She received her Ph.D. degree in Mechatronics Engineering from the GUC in Dec. 2022 in the field of Cooperative Architecture for Transportation Systems. She received her M.Sc. degree in Mechatronics Engineering from the GUC in Nov. 2017 in the field of cooperative control of multi-robot systems. She is currently the director of the Cognitive Driving Research in Vehicular Systems (C-DRiVeS Lab), which explores autonomous driving stack modules with a special focus on driving behavior in Egyptian traffic dynamics. The lab develops and curates region-specific datasets to capture the complexity of local road environments, aiming to build perception and decision-making models that reflect real-world driving conditions in Egypt and comparable emerging regions. She serves as a Board of Governors (BoG) member in the IEEE ITS Society during the interval 2023-2025, the 2023-2025 Co-chair of the committee on Diversity, Equity, and Inclusion in ITS committee chair. 
\endbio

\bio{bio_shehata.jpg}
Omar M. Shehata is an Associate Professor of Mechatronics and Robotics Engineering in the German University in Cairo (GUC), Egypt. He obtained his BSc in Mechatronics Engineering in 2010 from Ain Shams University (ASU). He obtained his 1st MSc from ASU in 2014 in Multi-Vehicle Coordination in intersections, and another MSc in 2015 from GUC in micro-robots control. In 2018 he finished his PhD from the Mechatronics Engineering Dept, ASU. In 2019, he finished his MBA from the school of Business, specializing in Strategic Planning. He is also the director of the Multi-Robot Systems (MRS) Research Group focusing on addressing the challenges related to the coordination of multi-robot and multi-vehicular systems. MRS application areas extend across different fields including robotics manipulators, legged robots, connected vehicles including the aerial and ground. His research interests include robot cooperation, computer vision and optimization techniques. Also, he is the founder and Head of the DRIVe-IT Hub, which is the central hub bridging challenges between Africa \& MENA region countries and the international Intelligent Systems (ITS) community.\endbio

\bio{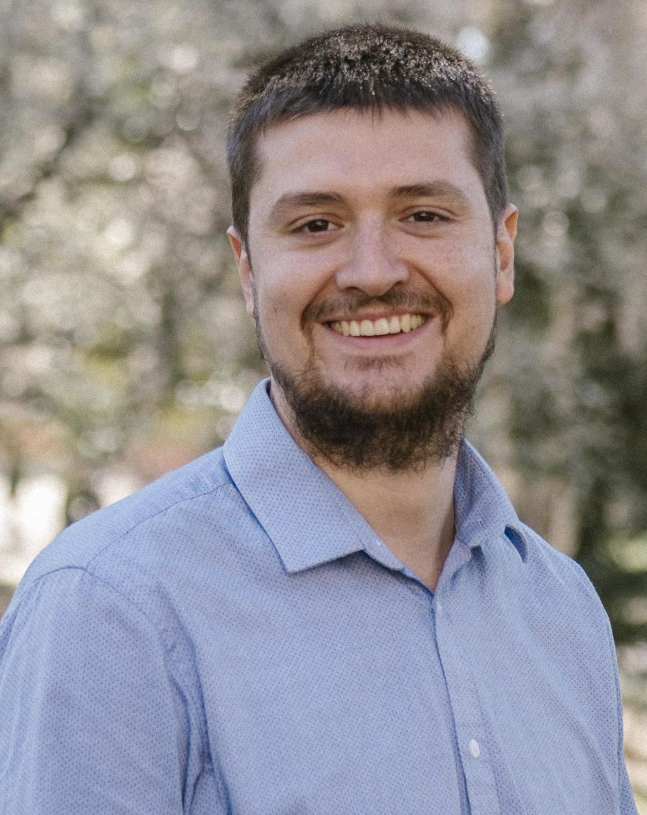}
Rubén Izquierdo received the Bachelor’s degree in electronics and industrial automation engineering in 2014, the M.S. in industrial engineering in 2016, and the Ph.D. degree in information and communication technologies in 2020 from the University of Alcalá (UAH). He is currently Assistant Professor at the Department of Computer Engineering of the UAH. His research interest is focused on the prediction of vehicle behaviors and control algorithms for highly automated and cooperative vehicles. His work has developed a predictive ACC and AES system for cut-in collision avoidance successfully tested in Euro NCAP tests. He was awarded with the Best Ph.D. thesis on Intelligent Transportation Systems by the Spanish Chapter of the ITSS in 2022, the outstanding award for his Ph.D. thesis by the UAH in 2021. He also received the XIII Prize from the Social Council of the UAH to the University-Society Knowledge Transfer in 2018 and the Prize to the Best Team with Full Automation in GCDC 2016.
\endbio

\bio{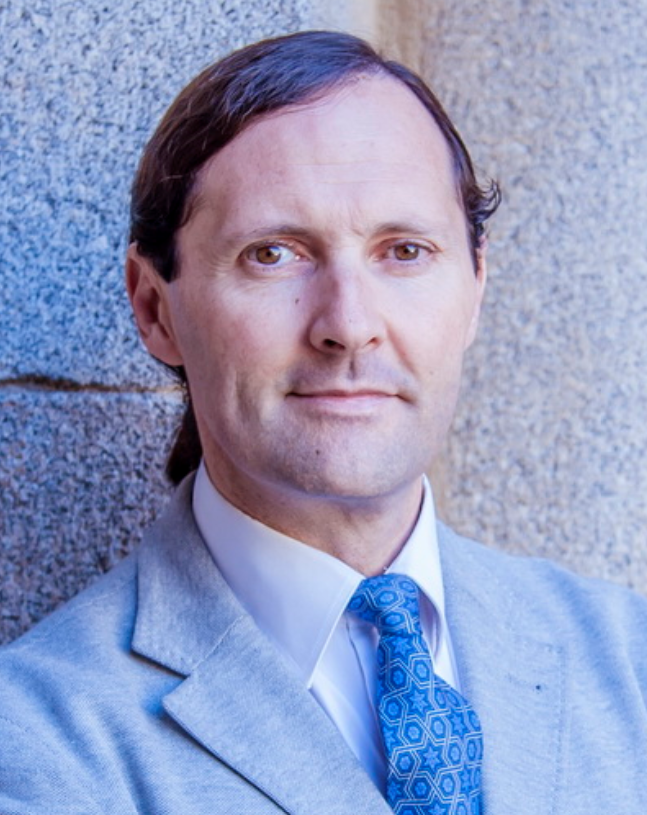}
Miguel \'Angel Sotelo received the degree in Electrical Engineering in 1996 from the Technical University of Madrid, the Ph.D. degree in Electrical Engineering in 2001 from the University of Alcalá (Alcalá de Henares, Madrid), Spain, and the Master in Business Administration (MBA) from the European Business School in 2008. He is currently a Full Professor at the Department of Computer Engineering of the University of Alcalá (UAH). His research interests include Self-driving cars, Prediction Systems, and Traffic Technologies. He is author of more than 300 publications in journals, conferences, and book chapters. He has been recipient of the Best Research Award in the domain of Automotive and Vehicle Applications in Spain in 2002 and 2009, and the 3M Foundation Awards in the category of eSafety in 2004 and 2009. Miguel Ángel Sotelo has served as Project Evaluator, Rapporteur, and Reviewer for the European Commission in the field of ICT for Intelligent Vehicles and Cooperative Systems in FP6 and FP7. He was Editor-in-Chief of the IEEE Intelligent Transportation Systems Magazine (2014-2016), Associate Editor of IEEE Transactions on Intelligent Transportation Systems (2008-2014), member of the Steering Committee of the IEEE Transactions on Intelligent Vehicles (since 2015), and a member of the Editorial Board of The Open Transportation Journal (2006-2015). He has served as General Chair of the 2012 IEEE Intelligent Vehicles Symposium (IV’2012) that was held in Alcalá de Henares (Spain) in June 2012. He was recipient of the IEEE ITS Outstanding Research Award in 2022, the IEEE ITS Outstanding Application Award in 2013, and the Prize to the Best Team with Full Automation in GCDC 2016. He is a Former President of the IEEE Intelligent Transportation Systems Society.
\endbio

\end{document}